\documentclass[sigconf,authorversion,nonacm]{acmart}

\usepackage{nicefrac}
\usepackage[utf8]{inputenc}
\usepackage{graphicx}
\usepackage{array,booktabs}
\usepackage{tabularx}
\usepackage{lscape}
\usepackage{tabularray}
\usepackage{xcolor}
\usepackage{makecell}
\usepackage{multirow}
\usepackage{tikz}
\usetikzlibrary{arrows,shapes}
\usetikzlibrary{arrows.meta}
\usepackage{dsfont}
\usepackage{amsmath, amsthm}
\theoremstyle{definition}
\newtheorem{definition}{Definition}
\usepackage{enumitem}
\usepackage{float}
\usepackage{lscape}
\usepackage{subcaption}

\newtheoremstyle{noparentheses}
  {\topsep}   
  {\topsep}   
  {\itshape}  
  {0pt}       
  {\bfseries} 
  {.}         
  {5pt plus 1pt minus 1pt} 
  {\thmname{#1} \thmnumber{#2} \thmnote{#3}}          
\theoremstyle{noparentheses}

\newtheorem{thm}{Theorem}[section]

\usepackage[createShortEnv]{proof-at-the-end}

\newtheorem{assumption}{Assumption}[section]
\newtheorem{lemma}{Lemma}[section]

\newcolumntype{s}{>{\arraybackslash}m{0.001cm}}

\usepackage{booktabs}
\newcommand{\tabitem}{~~\llap{\textbullet}~~}
\usepackage{thmtools} 
\usepackage{thm-restate}

\def\BibTeX{{\rm B\kern-.05em{\sc i\kern-.025em b}\kern-.08em
    T\kern-.1667em\lower.7ex\hbox{E}\kern-.125emX}}

\begin{document}

\title{A Systematic and Formal Study of the Impact of Local Differential Privacy on Fairness: Preliminary Results}

\author{Karima Makhlouf}
\email{karima.makhlouf@lix.polytechnique.fr}
\orcid{0000-0001-6318-0713}
\affiliation{%
  \institution{INRIA, École Polytechnique, IPP}
  \city{Paris}
  \country{France}
}

\author{Tamara Stefanovi\'{c}}
\email{tamara.stefanovic@mi.sanu.ac.rs}
\orcid{}
\affiliation{%
  \institution{Mathematical Institute of the Serbian Academy of Sciences and Arts}
  \city{Belgrade}
  \country{Serbia}
}

\author{Héber H. Arcolezi}
\email{heber.hwang-arcolezi@inria.fr}
\orcid{}
\affiliation{%
  \institution{Inria Centre at the University Grenoble Alpes}
  \city{Grenoble}
  \country{France}
}

\author{Catuscia Palamidessi}
\email{catuscia@lix.polytechnique.fr}
\orcid{0000-0003-4597-7002}
\affiliation{%
  \institution{INRIA, École Polytechnique, IPP}
  \city{Paris}
  \country{France}
}

\renewcommand{\shortauthors}{Makhlouf, et al.}

\begin{abstract}
Machine learning (ML) algorithms rely primarily on the availability of training data, and, depending on the domain, these data may include sensitive information about the data providers, thus leading to significant privacy issues. Differential privacy (DP) is the predominant solution for privacy-preserving ML, and the local model of DP is the preferred choice when the server or the data collector are not trusted. Recent experimental studies have shown that local DP can impact ML prediction for different subgroups of individuals, thus affecting fair decision-making. However, the results are conflicting in the sense that some studies show a positive impact of privacy on fairness while others show a negative one. In this work, we conduct a systematic and formal study of the effect of local DP on fairness. Specifically, we perform a quantitative study of how the fairness of the decisions made by the ML model changes under local DP for different levels of privacy and data distributions. In particular, we provide bounds in terms of the joint distributions and the privacy level, delimiting the extent to which local DP can impact the fairness of the model. We characterize the cases in which privacy reduces discrimination and those with the opposite effect. We validate our theoretical findings on synthetic and real-world datasets.
Our results are preliminary in the sense that, for now, we study only the case of one sensitive attribute, and only statistical disparity, conditional statistical disparity, and equal opportunity difference.
\end{abstract}


\maketitle

\pagestyle{plain}

\section{Introduction}
\label{sec:intro}
Information gathered about individuals is frequently utilized to make decisions that affect those same individuals.
For example,  financial services firms use ML models for risk management and real-time market data analysis. Medical researchers also use machine learning (ML) to identify disease signs and risk factors, and doctors to help diagnose illnesses and medical conditions in patients. 
In these contexts, there is a tension between the requirement for accurate systems ensuring individuals receive their due and the necessity to safeguard individuals from the improper exposure of their confidential information.
In other words, can these models be trusted to operate on personal and sensitive data? Are these models fair or do they potentially reproduce or exacerbate existing bias in society?

Differential privacy (DP)~\cite{dwork2016calibrating} is currently the leading privacy-preserving ML solution to protect sensitive information about individuals in data set used for statistical purposes or for training machine learning models. This is achieved by injecting controlled noise on the aggregated data or during the learning process. However, the original model of DP (aka central DP) assumes trustworthy data collectors and servers. For this reason, in recent years the local model of  DP (LDP)~\cite{kasiviswanathan2011can} has gained more and more attention, as it achieves privacy guarantees without the above assumptions. 
Indeed, with LDP each data point is locally obfuscated at the data-owner side before being collected, thus protecting data from privacy leaks at both the source and the server side. 
LDP has been endorsed and deployed by big tech companies. For instance, Google Chrome uses LDP to collect data from users~\cite{rappor}, and Apple uses LDP to collect emoji usage data, word usage, and other information from iPhone users (iOS keyboard)~\cite{apple_2017}.

On the other hand, algorithmic fairness strives to guarantee that generated models refrain from discriminating against groups or individuals on the basis of their sensitive attributes (such as race, gender, age, etc.). Numerous fairness metrics have been formally established and suggested in the literature to evaluate or quantify discrimination~\cite{10.1145/3468507.3468511}. These metrics can be broadly categorized into two main groups: group metrics and individual metrics. Group fairness metrics seek to guarantee uniform decisions across sub-populations, while individual fairness metrics aim to ensure that comparable individuals are treated equally~\cite{Makhlouf2021b, alves2022survey, mitchell2021algorithmic, 10.1145/3457607}.   

Achieving both privacy and fairness is crucial. However, it has been shown that privacy-preserving algorithms, and in particular (central and local) DP, tend to affect majority and minority groups differently, thus implying that in some cases privacy and fairness are at odds~\cite{farrand2020neither, 
 bagdasaryan2019differential, agarwal2021trade, chang2021privacy, pujol2020fair}. Nevertheless, in other lines of research, DP and fairness results align. For instance, Dwork et al.~\cite{dwork2012fairness} proved that individual fairness is a generalization of DP and provided some constraints under which a DP mechanism also ensures individual fairness. Xu et al.~\cite{xu2019achieving} proposed algorithms to achieve both DP and fairness in logistic regression by incorporating fairness constraints in the objective function. Recently, Arcolezi et al.~\cite{arcolezi2023local} proposed a novel privacy budget allocation scheme that considers the varying domain size of sensitive attributes and showed that, under this scheme, LDP leads to slightly improved fairness in learning tasks. \emph{These contrasting claims, most of which are backed only by experimental results, show that a systematic and foundational study of the relationship between privacy and fairness is highly needed}. This work is a step in that direction. 


Specifically, we formally study the impact of training a model with data obfuscated by randomized response (RR)~\cite{Warner1965}, a fundamental LDP protocol~\cite{kairouz2016discrete} that serves as a building block for more complex LDP mechanisms (e.g.,~\cite{Bassily2015,tianhao2017,rappor}).
The choice of RR is also motivated by its optimality for distribution estimation under several information theoretic utility functions~\cite{kairouz2016extremal} and by its design simplicity as it does not require any particular encoding. Specifically, RR provides optimal computational and communication costs for users since the output space equals the input space. Moreover, no decoding step is needed on the server side. It also means that the server is free to use any post-processing coding techniques (e.g., one-hot encoding, mean encoding, binary encoding) to improve the usefulness of the ML model.

Building on this foundation, our main contribution consists of a theoretical analysis of how the fairness of the prediction of an ML model is affected by the application of RR on the training data, depending on the level of privacy 
and the data distribution. In particular, we study three notions of fairness: statistical disparity~\cite{dwork2012fairness}, conditional statistical disparity~\cite{corbett2017algorithmic}, and equal opportunity~\cite{hardt2016equality}, and identify the conditions under which they are improved or reduced by RR. We then empirically validate our results by performing experiments on synthetic data and four real datasets, \textit{Compas}~\cite{angwin2016machine}, \textit{Adult}~\cite{ding2021retiring}, \textit{German credit}~\cite{Dua:2019}, and \textit{LSAC}~\cite{wightman1998lsac}. All detailed proofs supporting our findings are available in Appendix~\ref{sec:proofs}.
 


\section{Related Work}
\label{sec:related}
Although the topics of privacy and fairness have been studied for years by philosophers and sociologists, they have only recently received significant attention from the computer science community. This section provides an overview of existing research that explores the intriguing relationship between DP and fairness, specifically in the context of ML. We also recommend the recent survey by Fioretto et al.~\cite{fioretto2022differential}, which discusses the conditions under which DP and fairness have aligned or contrasting goals in decision and learning tasks.
 

{\bf Central differential privacy.} 
Pujol et al.~\cite{pujol2020fair} empirically measure the impact of differentially private algorithms on allocation processes. They use two privacy mechanisms: the Laplace and the Data-and-Workload-Aware algorithm. Their results show that in the settings where the introduced noise is modest (higher $\varepsilon$), impacts on fairness may be negligible. However, the introduced noise disproportionately impacts different groups under strict privacy constraints (smaller $\varepsilon$). In~\cite{bagdasaryan2019differential}, the authors empirically showed that by introducing noise, the accuracy of a model trained using DP-SGD~\cite{abadi2016deep} decreases compared to the original, non-private model. More specifically, if the original model is “unfair” (in the sense that accuracy is not the same across different subgroups), then DP-SGD deepens the differences between subgroups. 
Mangold et al. in~\cite{mangold2023differential} performs a theoretical analysis of the impact of central DP on fairness in classification. They prove that the difference in fairness levels between private and non-private models diminishes at a rate of $\Tilde{O}\left(\sqrt{p}/n\right)$, where $n$ represents the number of training records and $p$ is the number of parameters. They also provide an empirical study using the central model with Gaussian noise for DP and $l2$-regularized logistic regression models for prediction. 

{\bf Local differential privacy.} In~\cite{mozannar2020fair}, Mozannar et al. show how to adapt non-discriminatory learners to work with privatized attributes, giving theoretical guarantees on performance. The experimental analysis by Makhlouf et al.~\cite{makhlouf2023impact} showed that obfuscating several sensitive attributes instead of obfuscating only the sensitive attribute used to assess fairness gives better results for fairness. Also, the authors observed that combined LDP, compared to independent LDP, reduces the disparity more efficiently at low privacy guarantees (high $\varepsilon$). 
Arcolezi et al.~\cite{arcolezi2023local} also empirically deals with the impact on fairness of applying  LDP to multiple sensitive attributes. The analysis covers several fairness metrics and state-of-the-art LDP protocols. Their results contrast with those obtained with central DP, as they show that LDP slightly improves fairness in learning tasks without significant loss of the accuracy of the model.





\section{Preliminaries and Notation}
\label{sec:preliminaries}
This section presents the framework we consider in this work and briefly recalls the privacy setting and the fairness metrics applied in this study. 

Variables are denoted by uppercase letters, while lowercase letters denote specific values of variables (e.g., $A=a$, $Y=y$).
A predictor $\hat{Y}$ of an outcome $Y$ is a function of a set of variables $(A,X)$ where $X$
designates the set of non-sensitive attributes and $A \in \{0,1\}$ represents the sensitive attribute\footnote{In this work, we consider a single sensitive attribute (no intersectionality~\cite{Makhlouf2021b}) and $X$ can be a vector of variables.}. For example, when deciding to hire an individual, the sensitive attribute could be someone's gender or race, and the non-sensitive attributes $X$ could include the person's education level and professional experience. Note that $X$ could include proxies to $A$, such as zip code, which could hint to race. We assume that $\hat{Y}$ and $Y$ are binary random variables where $Y = 1$ (e.g., hiring a person) designates a positive outcome, and $Y = 0$ (e.g., not hiring a person) designates a negative outcome.  
For the remainder of this paper, we assume that we have access to a (multi)set $S = \{ (a_i,x_i,y_i) \}_{i=1}^{n}$ of $n$ i.i.d samples from the distribution on $A\times X \times Y$. 

We call $A'= \mathcal{L}(A)$ the obfuscated version of the sensitive attribute $A$, where $\mathcal{L}$ is a certain randomized LDP mechanism. Thus, we denote a randomized version of $S$ as $S'= {(a'_i,x_i,y_i)}_{i=1}^{n}$. 

\subsection{The framework}
\label{subsec:framework}
\begin{figure*}[!ht]
\centering
    \includegraphics[scale=0.45]{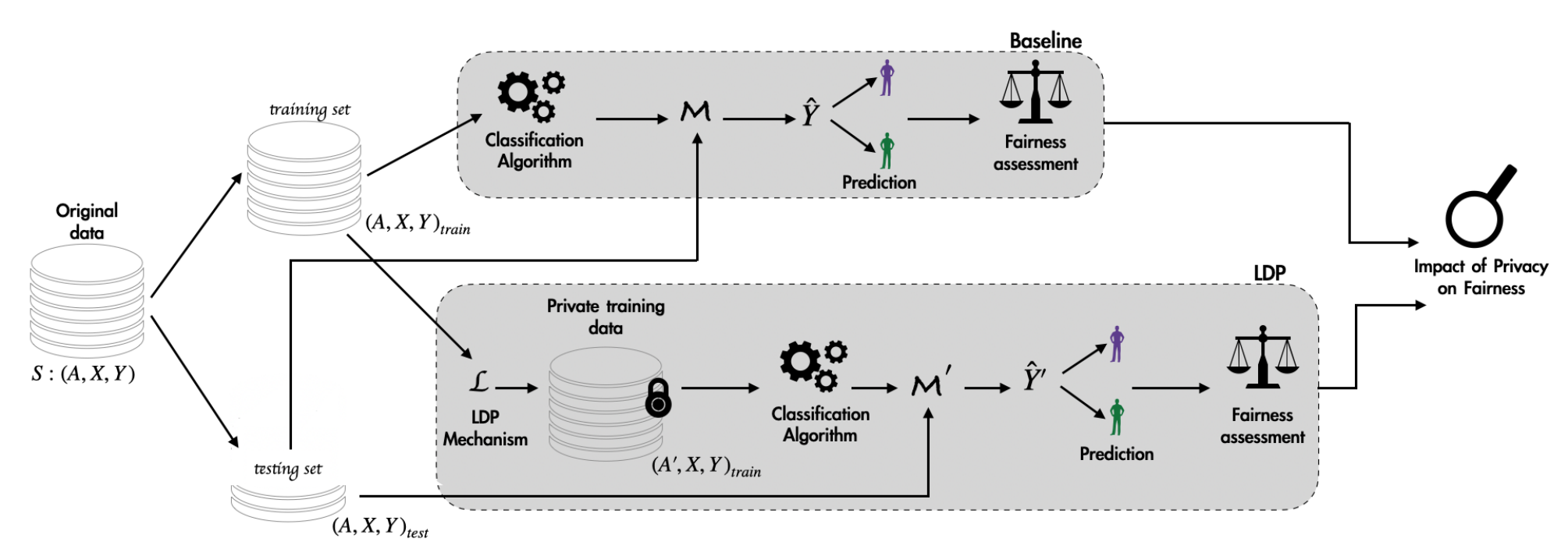} 
    \caption{Our framework to assess the impact of LDP on the fairness of a ML model.}
    \label{fig:framework}
\end{figure*}

Figure~\ref{fig:framework} illustrates the framework deployed in our work.
We assume a given decision task, such as deciding whether to release a convict on parole or admit an applicant to a college program. We assume that we dispose of a set of data $S = (A,X,Y)_{\mathit train}\;
\bigcup \; (A,X,Y)_{\mathit test}$ 
for building an ML model to help with the task, and for evaluating it. Specifically, $(A,X,Y)_{\mathit train}$ is used for training the model, and 
$(A,X,Y)_{\mathit test}$ to assess the fairness of its predictions. 

As shown in Figure~\ref{fig:framework}, in order to measure the impact of the LDP mechanism $\mathcal{L}$, we train two models. The baseline model  $\mathcal{M}$  (upper shaded box) is trained on the original data $(A,X,Y)_{\mathit train}$, and we call its prediction $\hat{Y}$.
Then, we obfuscate the training set by applying  $\mathcal{L}$ to the $A$ component of each sample in $(A,X,Y)_{train}$. The resulting data set $(A',X,Y)_{train}$ is used to train (with the same classification algorithm and  the same hyper-parameters)
a second model $\mathcal{M}'$, whose prediction is called  $\hat{Y'}$ (lower shaded box). 

The difference between $\hat{Y'}$ and $\hat{Y}$ on the original testing data quantifies the impact of LDP on the fairness of the model.
\emph{It is important to emphasize that, in our framework,  the individual predictions, both for $\mathcal{M}$ and $\mathcal{M}'$,  are 
obtained by applying the models to the original testing data $(A,X,Y)_{\mathit test}$}. Namely, in testing phase, $\hat{Y}= \mathcal{M}(A,X)$ and $\hat{Y}'= \mathcal{M}(A,X)$ (instead of $\hat{Y}'= \mathcal{M}(A',X)$). 
This is because we argue that fairness must be evaluated on the true data. Indeed,  even if a model was trained on obfuscated data, it is likely to receive the true data as input at the moment of its deployment. And in any case, the presence of proxies may reveal the true value of the sensitive variable anyway.

\subsection{Local Differential Privacy}
\label{subsec:privacy}

We recall the definition of LDP as given in the literature for the discrete case. 
\begin{definition}[$\varepsilon$-Local Differential Privacy~\cite{kairouz2016discrete}]\label{def:ldp} An algorithm $\mathcal{L}$ satisfies $\varepsilon$-LDP, where $\varepsilon$ is a positive real number representing the privacy parameter, if for any input $v_1$ and $v_2 \in Dom(\mathcal{L})$ and for all possible output $y$:

\begin{equation*} \label{eq:ldp}
    \mathbb{P}[\mathcal{L}(v_1) = y] \leq e^\varepsilon \; \mathbb{P}[\mathcal{L}(v_2) = y]  
\end{equation*}
\end{definition}


Several LDP mechanisms have been proposed in the literature~\cite{Xiong2020}. The mechanism we consider here is randomized response (RR)~\cite{Warner1965,kairouz2016discrete} for a binary variable $a\in \{0,1\}$, which is defined in Equation~\eqref{eq:RR}. 

 \begin{equation} \label{eq:RR}
    {\rm RR}(a) = \begin{cases} a \quad \textrm{with probability} \quad \frac{e^{\varepsilon}}{e^{\varepsilon}+1}  \\ 
    \overline{a} \quad \textrm{with probability} \quad \frac{1}{e^{\varepsilon}+1}  \textrm{.}\end{cases} 
    \end{equation}

\noindent where $\overline{a}=1$ if  $a=0$ and, viceversa, $\overline{a}=0$ if  $a=1$. It is easy to see that RR satisfies $\varepsilon$-LDP.
For simplicity, we will denote by   $p$ the probability  $\nicefrac{e^{\varepsilon}}{(e^{\varepsilon}+1)}$
that the reported value is the true value.


\subsection{Fairness}
\label{subsec:fairness}
Many fairness metrics have been proposed in the literature, and they fall into two main categories, namely, the group and the individual fairness metrics~\cite{10.1145/3457607, 10.1145/3468507.3468511, verma2018fairness,barocas-hardt-narayanan,Makhlouf2021b, alves2022survey}.
This paper focuses on statistical group fairness metrics, which assess fairness based on a predefined sensitive attribute $A$. 
We will call \emph{privileged} the group  for which the prediction is favorable ($\hat{Y}=1$) more frequently, and 
\emph{unprivileged} the other group.
We will consider the following metrics.

\begin{itemize}[leftmargin=*]
    \item \textbf{Statistical disparity (SD)}~\cite{dwork2012fairness}
    is the most basic notion of fairness.
    It measures the difference in acceptance rates between groups and is defined as:
    \begin{equation}
    \small
\label{eq:sd}
\mbox{SD} = \mathbb{P}[\hat{Y}=1 \mid A = 1] - \mathbb{P}[\hat{Y}=1 \mid A = 0].
\end{equation}
\item \textbf{Conditional statistical disparity (CSD)}~\cite{corbett2017algorithmic} is a variant of statistical disparity obtained by conditioning on a set of explanatory attributes which 
may legitimate the discrimination~\cite{kamiran2013quantifying}.
In this paper, we assume that all variables in $X$ are (potential) 
explanatory variables, and 
we define conditional statistical  disparity for each instance $x$ of $X$ as follows:
\begin{equation}
\small 
\label{eq:csd}
\mbox{CSD}_x = \mathbb{P}[\hat{Y}=1 \mid X=x,A = 1] - \mathbb{P}[\hat{Y}=1 \mid X=x,A = 0]. 
\end{equation}
Note that, in general, $x$ represents a tuple of values since $X$ may contain more than one attribute. 
\item \textbf{Equal opportunity difference (EOD)}~\cite{hardt2016equality} is one of the most popular notions of fairness nowadays. It measures the disparity between the  true positive rate\footnote{The true positive rate is defined as $\frac{TP}{TP+FN}$, where $TP$ are the true positive predictions and $FN$ are the false negative predictions.} among the two groups:
\begin{equation}
\small
\label{eq:EOD}
    \mbox{EOD} = \mathbb{P}[\hat{Y}=1 \mid Y=1,A = 1] - \mathbb{P}[\hat{Y}=1\mid Y=1,A = 0].
\end{equation}
\end{itemize}

\section{Quantitative analysis of the Impact of Privacy on Fairness}
\label{sec:quantification}

In this section, we formally study the impact of LDP on fairness. Specifically, we perform a quantitative study of how the fairness of the prediction is affected by the application of the  RR mechanism to the sensitive values in the training data, depending on the level $\varepsilon$ of privacy and on the data distribution. 

We briefly recall our setting. In addition to the sensitive attribute $A$ and the true decision $Y$, which are binary, the data includes a set of non-sensitive attributes $X$ with arbitrary values. We assume that the data model is \emph{probabilistic}, 
in the sense that the data may contain tuples with the same values for $X$ and $A$ and different values for $Y$. 
 $A'={\rm RR}(A)$ is an obfuscated\footnote{In this paper, we use the terms \textit{obfuscated} and \textit{randomized} exchangeably.} version of $A$ obtained by applying the RR mechanism to $A$, and it is also binary. The prediction of the model trained on the original data is denoted by $\hat{Y}$, while that of the model trained on the obfuscated data, which we will call LDP model, is  $\hat{Y}'$. Of course, $\hat{Y}$ and $\hat{Y}'$ are also binary. We assume that both models are deterministic. Namely, on a given input $(x,a)$, ${\mathcal M}$ always outputs the same prediction. 
 The same holds for ${\mathcal M}'$, although the prediction  may be different from the one of 
 ${\mathcal M}$.

Table~\ref{tab:def} shows some abbreviations and definitions we use in the paper. In particular, $\Delta_a^x$ denotes the difference between the frequency of the samples with the positive true decision ($Y=1$) and those with the negative true decision ($Y=0$), and have $A=a$ and $X=x$. On the other hand, $\Gamma_a^x$ denotes the difference between the positive and negative decision rates \emph{given} $A=a$ and $X=x$. $\Delta_a'^x$ and $\Gamma_a'^x$ denote the corresponding quantities in the obfuscated training data (i.e., on the samples with $A'=a$ and $X=x$). 


\begin{table*}[ht]
 \centering
\caption{Abbreviations and definitions used in the paper. $\mathbb{\hat{P}}$ denotes the empirical probability (frequency) on the training set.}
\label{tab:def}
\renewcommand\arraystretch{1.2}
\begin{tabular}{@{} *1l @{}}\toprule
\emph{Abbreviations} \\ \midrule
\vspace{1.2mm}
\tabitem $\hat{Y}^x_a\in\{0,1\}$ : the prediction of the baseline model $\mathcal M$ on the input $(x,a)$
\\
\tabitem $\mbox{CSD}_x = \mathbb{P}[\hat{Y}=1 \mid X= x, A = 1] - \mathbb{P}[\hat{Y}=1 \mid X=x, A = 0] $ : Conditional statistical disparity in $\mathcal M$
\\[.8mm]
\tabitem ${\rm SD} = \mathbb{P}[\hat{Y}=1 \mid A = 1] - \mathbb{P}[\hat{Y}=1 \mid A = 0]$ : statistical disparity in $\mathcal M$
\\[.8mm]
\tabitem
${\rm EOD} = \mathbb{P}[\hat{Y}=1 \mid Y=1, A = 1] - \mathbb{P}[\hat{Y}=1 \mid Y=1, A = 0]$ : equal opportunity difference in $\mathcal M$ 
\\[.8mm]
\midrule
\tabitem $\hat{Y}'^x_a\in\{0,1\}$ : the prediction of the LDP model ${\mathcal M}'$ on the input $(x,a)$
\\[.8mm]
\tabitem $\mbox{CSD}'_x  = \mathbb{P}[\hat{Y}'=1 \mid X= x, A = 1] - \mathbb{P}[\hat{Y}'=1 \mid X=x, A = 0] $
 : Conditional statistical disparity in ${\mathcal M}'$
\\[.8mm]
\tabitem ${\rm SD}' = \mathbb{P}[\hat{Y}'=1 \mid A = 1] - \mathbb{P}[\hat{Y}'=1 \mid A = 0]$ : statistical disparity in ${\mathcal M}'$
\\[.8mm]
\tabitem
${\rm EOD}' = \mathbb{P}[\hat{Y}'=1 \mid Y=1, A = 1] - \mathbb{P}[\hat{Y}'=1 \mid Y=1, A = 0]$ : equal opportunity difference in ${\mathcal M}'$
\\[1.2mm]
\hline
\hline
\emph{Definitions} \\ 
\midrule
\vspace{1.2mm}
\tabitem $\Delta_a^x = \mathbb{\hat{P}}[Y=1,X=x,A=a]-\mathbb{\hat{P}}[Y=0,X=x,A=a]$ \\ \vspace{1.2mm}
\tabitem $\Gamma_a^x = \mathbb{\hat{P}}[Y=1|X=x,A=a]-\mathbb{\hat{P}}[Y=0|X=x,A=a]$ 
\\ 
\midrule
\vspace{2mm}
\tabitem $\Delta_a'^x = \mathbb{\hat{P}}[Y=1,X=x,A'=a]-\mathbb{\hat{P}}[Y=0,X=x,A'=a]$ \\ \vspace{1.2mm}
\tabitem $\Gamma_a'^x = \mathbb{\hat{P}}[Y=1|X=x,A'=a]-\mathbb{\hat{P}}[Y=0|X=x,A'=a]$ \\


\bottomrule
\hline
\end{tabular}
\end{table*}

In order to reason formally about the impact of privacy on fairness, we need to make a basic assumption about the training algorithm. Namely, we assume that the baseline model, in correspondence of the input $(x,a)$, predicts $\hat{Y}=1$ if $\Delta^x_a \geq 0$, namely 
the majority of the tuples in the training set with $X=x$ and $A=a$ have $Y=1$, and predicts $\hat{Y}=0$, otherwise. This assumption is quite natural, as, in general, an ML model should opt for the prevailing decision seen in training\footnote{Some learning algorithms like the Nearest Neighbours actually use a generalization of this criterion to produce the prediction.}. 
We make the same assumption for the LDP model $\mathcal{M'}$ (with $A$ replaced by $A'$), which is reasonable since $\mathcal{M}$ and $\mathcal{M'}$ are trained with the same algorithm. Formally:

\begin{assumption}\label{assumption1}
The prediction of $\mathcal{M}$ (baseline model) is:
\begin{equation}\label{eq:baseline_model}
   \hat{Y}^x_a \,=\, \begin{cases} 1 \quad \textrm{if} \quad \Delta^x_a \geq 0 \quad  (\textrm{or, equivalently,}\,\, \Gamma_a^x \geq 0) \textrm{,}\\ 
    0 \quad \textrm{otherwise.}
    \end{cases} \nonumber 
\end{equation}
\end{assumption}

\begin{assumption}\label{assumption2}
The prediction of $\mathcal{M'}$ (LDP model) is:
\begin{equation}\label{eq:ldp_model}
    \hat{Y}_a'^x \,= \,\begin{cases} 1 \quad \textrm{if} \quad \Delta_a'^x \geq 0 \quad  (\textrm{or, equivalently,} \,\,\Gamma_a'^x \geq 0) \textrm{,}\\ 
        0 \quad \textrm{otherwise.}
        \end{cases} \nonumber
\end{equation}
\end{assumption}

The following Lemma relates the difference 
between the frequencies of positive and negative decisions in the obfuscated and original data. We recall that $p=\nicefrac{e^{\varepsilon}}{(e^{\varepsilon}+1)}$ is the probability that the value reported by RR is the true value. 

\begin{lemmaE}[][end,restate]
\label{lemma1}
   $\Delta_a'^x = p \; \Delta^x_a + (1-p) \; \Delta^x_{\overline{a}}$ .
\end{lemmaE}

\begin{proofE}
\begin{align} 
    \Delta_a'^x = \; &  \mathbb{\hat{P}}[Y=1,X=x,A'=a] - \mathbb{\hat{P}}[Y=0,X=x,A'=a]\nonumber \\
    =\; & p \; \mathbb{\hat{P}}[Y=1,X=x,A=a] + (1-p)  \mathbb{\hat{P}}[Y=1,X=x,A=\overline{a}] - \left ( p \; \mathbb{\hat{P}}[Y=0,X=x,A=a] + (1-p)  \mathbb{\hat{P}}[Y=0,X=x,A=\overline{a}] \right )  \nonumber \\
    = \; & p \left ( \mathbb{\hat{P}}[Y=1,X=x,A=a] - \mathbb{\hat{P}}[Y=0,X=x,A=a]    \right )  + (1-p) \left ( \mathbb{\hat{P}}[Y=1,X=x,A=\overline{a}] - \mathbb{\hat{P}}[Y=0,X=x,A=\overline{a}]    \right )  \nonumber \\
    = \; &  p \; \Delta^x_a + (1-p) \; \Delta^x_{\overline{a}} \nonumber
\end{align}
\end{proofE}

The following Lemma relates the LDP model's prediction to the original data's statistics. It follows simply by case analysis from Lemma~\ref{lemma1} and Assumption~\ref{assumption2}.

\begin{lemma}~\label{lemma2}
    \begin{equation}
    \footnotesize
    \begin{array}{l}
   \hat{Y}'^x_a \,=  \, 1 \quad\textrm{if} \quad \begin{cases} 
   \, \Delta^x_a, \, \Delta^x_{\overline{a}} \geq 0, \;\; \textrm{or} \\ 
   \vspace{0.1cm}
  \, \Delta^x_a > 0 \quad\textrm{and} \quad\Delta^x_{\overline{a}} < 0 \quad \textrm{and } \quad e^\varepsilon \geq 
- \, \nicefrac{\Delta^x_{\overline{a}}}{\Delta^x_a}, \;\; \textrm{or} \\
\, \Delta^x_a < 0 \quad\textrm{and} \quad\Delta^x_{\overline{a}} > 0 \quad\textrm{and } \quad e^\varepsilon \leq 
- \,\nicefrac{\Delta^x_{\overline{a}}}{\Delta^x_a} \, . \\
    \end{cases} \nonumber 
    \\ \ \\
   \hat{Y}'^x_a \,= \, 0 \quad\textrm{if} \quad \begin{cases} 
   \, \Delta^x_a, \,\Delta^x_{\overline{a}} \leq 0  \quad \textrm{and at least one of them is strictly negative,\,\, or} \\
  \vspace{0.1cm}
  \, \Delta^x_a > 0 \quad\textrm{and} \quad\Delta^x_{\overline{a}} < 0 \quad \textrm{and } \quad e^\varepsilon < 
    - \, \nicefrac{\Delta^x_{\overline{a}}}{\Delta^x_a}, \;\; \textrm{or} \\
 \, \Delta^x_a < 0 \quad\textrm{and} \quad \Delta^x_{\overline{a}} > 0 \quad\textrm{and } \quad e^\varepsilon > 
    -\, \nicefrac{\Delta^x_{\overline{a}}}{\Delta^x_a} \, .
    \\ 
  \end{cases} \nonumber 
  \end{array}
    \end{equation}
\end{lemma}



\subsection{Impact of LDP on conditional statistical disparity} \label{sub-sec:ldp_csd}


In this section, we analyze the effect of RR on conditional statistical disparity with respect to a specific tuple of values $x$ of the explaining variables. 
To do so, we compare  CSD$'_x$, which represents the conditional statistical disparity of prediction of the LDP model,  with CSD$_x$ which is the one of the baseline model. Following the principle that fairness should be assessed on the true inputs, we define CSD$'_x$ as:
\small 
\[ 
\mbox{CSD}'_x \, = \, \mathbb{P}[\hat{Y}'=1 \mid X= x, A = 1] - \mathbb{P}[\hat{Y}'=1 \mid X=x, A = 0].  \]
Namely, the conditioning is on $A$ and not on $A'$.
Note that, since the models are deterministic, CSD$_x$  and CSD$'_x$ could equivalently be defined as:
\[
\mbox{CSD}_x \, = \, \hat{Y}^x_1 - \hat{Y}^x_0
 \qquad \mbox{and} \qquad \mbox{CSD}'_x \, = \, \hat{Y}'^x_1 - \hat{Y}'^x_0\ .
\]

The following theorem states the relation between ${\rm CSD}_x$  and ${\rm CSD}'_x$.
\begin{thmE}[Impact of LDP on CSD$_x$][end, restate]\ \\
\label{th:csd}
\begin{enumerate}
    \vspace{-0.1cm}
    \item if \; ${\rm CSD}_x > 0$\; then \; $0 \leq {\rm CSD}'_x \leq {\rm CSD}_x$
    \vspace{0.2cm}
    \item if \; ${\rm CSD}_x < 0$\; then \; ${\rm CSD}_x \leq {\rm CSD}'_x \leq 0$
    \vspace{0.2cm}
    \item if \; ${\rm CSD}_x = 0$\; then \; ${\rm CSD}'_x =  {\rm CSD}_x = 0$
\end{enumerate}
\end{thmE}
\begin{proofE}
$ $\newline
  \begin{enumerate}
    \item if CSD$_x > 0$ then, according to Assumption~\ref{assumption1}, $\hat{Y}_1 = 1$ and $\hat{Y}_0 = 0$. \\
    Hence, $\Delta^x_1 \geq 0$ and $\Delta^x_0 <  0$. \\
    Using Lemma~\ref{lemma2}, we have:
    \begin{equation}
   \hat{Y}'^x_1 =   \begin{cases} 1 \;\; 
   \vspace{0.1cm}\textrm{if}\;\; \Delta^x_1 >0 \;\; \textrm{and } e^\varepsilon \geq 
- \nicefrac{\Delta^x_0}{\Delta^x_1}\\ 
   \vspace{0.1cm}
   0 \;\; \textrm{if}\;\; (\Delta^x_1 >0 \;\; \textrm{and }\;\; e^\varepsilon < 
- \nicefrac{\Delta^x_0}{\Delta^x_1})
\;\;  \textrm{or} \;\; \Delta^x_1  = 0 \\
    \end{cases} \nonumber 
    \end{equation} \\
    and \\
    \begin{equation}
   \hat{Y}'^x_0 =   \begin{cases} 1 \;\; 
   \vspace{0.1cm}\textrm{if}\;\; \Delta^x_1 >0 \;\; \textrm{and } e^\varepsilon \leq 
- \nicefrac{\Delta^x_1}{\Delta^x_0}\\ 
   \vspace{0.1cm}
   0 \;\; \textrm{if}\;\; (\Delta^x_1  > 0 \;\; \textrm{and } \;\; e^\varepsilon > 
- \nicefrac{\Delta^x_1}{\Delta^x_0})  \;\; \textrm{or} \;\;  \Delta^x_1 = 0 \\
    \end{cases} \nonumber 
    \end{equation} \\
Consequently, three scenarios are possible:\\
\begin{align}
    \bullet \; \hat{Y}'^x_1 = 0 \wedge \hat{Y}'^x_0 = 0 & \;\;  \textrm{if}  \;\; \Delta^x_1 =0 \nonumber \\
    & \;\; \textrm{or} \;\; \Delta^x_1  > 0 \;\;\textrm{and} \;\; e^\varepsilon < 
- \nicefrac{\Delta^x_0}{\Delta^x_1} \;\; \textrm{and} \;\; e^\varepsilon > 
- \nicefrac{\Delta^x_1}{\Delta^x_0} \nonumber \\
    \bullet \; \hat{Y}'^x_1 = 1 \wedge \hat{Y}'^x_0 = 0 & \;\;  \textrm{if}  \;\; \Delta^x_1  > 0 \;\;\textrm{and} \;\; e^\varepsilon \geq 
- \nicefrac{\Delta^x_0}{\Delta^x_1} \;\; \textrm{and} \;\; e^\varepsilon > 
- \nicefrac{\Delta^x_1}{\Delta^x_0} \nonumber \\
 \bullet \; \hat{Y}'^x_1 = 1 \wedge \hat{Y}'^x_0 = 1 & \;\;  \textrm{if}  \;\; \Delta^x_1  > 0 \;\;\textrm{and} \;\; e^\varepsilon \geq 
- \nicefrac{\Delta^x_0}{\Delta^x_1} \;\; \textrm{and} \;\; e^\varepsilon \leq 
- \nicefrac{\Delta^x_1}{\Delta^x_0}  \nonumber 
\end{align}

\noindent
Note that the case $\hat{Y}'^x_1 = 0 \; \wedge \; \hat{Y}'^x_0 = 1$ is not possible. Indeed, $\hat{Y}'^x_1 = 0 \; \wedge \; \hat{Y}'^x_0 = 1$ implies   $e^\varepsilon < 
- \nicefrac{\Delta^x_0}{\Delta^x_1}$ and $e^\varepsilon \leq 
- \nicefrac{\Delta^x_1}{\Delta^x_0}$. Note that  
the two fractions are one the inverse of the other. Hence, one of them is smaller than $1$, or both are $1$. Therefore, we would have 
$e^\varepsilon < 1$, which is not possible because $\varepsilon\geq 0$.  \\
\noindent 
Hence we have ${\rm CSD}'_x = 0$ or ${\rm CSD}'_x = 1$, i.e., $0\leq  {\rm CSD}'_x \leq {\rm CSD}_x$. 
\\
\item Case 2 (CSD$_x < 0$) is analogous to case 1. That is, proving this case amounts to replacing 0 by 1 and 1 by 0 in case 1 proof.  \\
\item if CSD$_x = 0$, two cases are possibles:
\begin{itemize}
 \item $ \hat{Y}^x_1 = 0 \wedge \hat{Y}^x_0 = 0$. This means that 
    $\Delta^x_1 < 0 \wedge \Delta^x_0 < 0$.  By Lemma~\ref{lemma2},  we  
    derive $\hat{Y}'^x_1 = 0 \wedge \hat{Y}'^x_0 = 0$. 
    Hence, CSD$'_x = 0$.
\\
\item $ \hat{Y}^x_1 = 1 \wedge \hat{Y}^x_0 = 1$. This means that 
    $\Delta^x_1 \geq 0 \wedge \Delta^x_0 \geq 0$.  By Lemma~\ref{lemma2},  we  
    derive $\hat{Y}'^x_1 = 1 \wedge \hat{Y}'^x_0 = 1$. 
    Hence, CSD$'_x = 0$.
\end{itemize}
 
\end{enumerate}
\end{proofE}

Essentially, the above theorem says that CSD$'_x$ is always sandwiched between  CSD$_x$ and $0$. 
Namely, if, in the baseline model, there is 
discrimination against one group, then obfuscating $A$ tends to reduce the discrimination. It never introduces discrimination against the other group. In one extreme case, it may leave things unchanged, while, in the opposite extreme case, it may remove the discrimination entirely.  
If, in the baseline model, we have conditional statistical parity (CSD$_x=0$), then obfuscating $A$ maintains this property. 

It is important to note that Theorem~\ref{th:csd} does not depend on whether the \textit{unprivileged} group is the minority or the majority of the population.  



\subsection{Impact of LDP on statistical disparity} \label{sub-sec:ldp_sd}
Using the results of CSD$_x$, in this section, we analyze the impact of privacy on SD by comparing SD$'$ and SD, where SD$'$ is the statistical disparity of the prediction of the LDP model, defined as: 
\begin{equation} \label{eq:sd'}
    \mbox{SD}' = \mathbb{P}[\hat{Y'}=1 \mid A = 1] - \mathbb{P}[\hat{Y'}=1 \mid A = 0].
\end{equation}
Again, note that we condition on $A$ rather than $A'$.

We make the following assumption that we call the \textit{uniform discrimination assumption}. Essentially, it says that if one group is discriminated against for some value $x*$ of $X$, then the other group cannot be discriminated against for other values $x$ of $X$. This is a natural assumption in real-life scenarios. For example, consider an ML system that tries to predict whether to release an individual on parole, given the type of crime they have committed in the past. If the system (or the historical data in which it is trained) discriminates against an ethnic group in case of a minor crime, it would still discriminate against that same group in case of a major crime, or, at most, be fair.
As another example, consider granting an application for a loan: If, for a certain amount of money requested, the applications from an ethnic group are accepted more frequently than those from the other group, it is unlikely that, for a different amount of money, the situation would be inverted.    

Formally, the \textit{uniform discrimination assumption} is stated as follows:

\begin{assumption}\label{assumption3} Uniform discrimination assumption
\begin{equation} \label{eq:uniformity}
     \text{if} \; \; \exists x^* \; \Gamma^{x^*}_a > \Gamma^{x^*}_{\overline{a}} \; \text{then} \; \; \forall x \; \; \Gamma ^x_a \geq \Gamma ^x_{\overline{a}}  \nonumber
\end{equation}
\end{assumption}

In the remainder of this section, we differentiate between two scenarios depending on whether $X$ and $A$ are independent. We will denote the case of independency by $X \perp A$, and the case of dependency by  $X \not \perp A$\footnote{In real-life contexts, $X$ and $A$ are usually dependent.}. 
\\
\noindent
\subsubsection{First scenario: $X \perp A$} \label{subsubsec:independent_xa}
\ \\
We first consider the case of independency. 
We start by showing that we can  quantitatively express SD in terms of the distribution of the data as follows:

\begin{lemmaE}[Quantification of SD][end, restate]
\label{lemma:SD}
\begin{align}
    {\rm SD} \, = \, \begin{cases} 
   \begin{array}{ll}
   \mathbb{P}[\Delta^{X}_1 \geq 0 \; \wedge \; \Delta^{X}_0 < 0] &  \textrm{if} \; \; \exists x \; \Gamma^{x}_1 > \Gamma^{x}_0 \\[2mm] 
   0 & \textrm{if} \; \; \forall x \; \Gamma^{x}_1 = \Gamma^{x}_0 \\[2mm]
    -  \; \mathbb{P}[\Delta^{X}_1 < 0 \; \wedge \; \Delta^{X}_0 \geq 0] &\textrm{if} \; \; \exists x \; \Gamma^{x}_1 < \Gamma^{x}_0 
    \end{array}
    \end{cases} \nonumber
\end{align}    
\end{lemmaE}
\begin{proofE}
\begin{align}
    {\rm SD}  & \stackrel{\rm def}{=} \mathbb{P}[\hat{Y}=1|A=1] - \mathbb{P}[\hat{Y}=1|A=0]  \nonumber \\
    & = \sum_{x} \mathbb{P}[\hat{Y}=1,X=x| A=1] - \sum_{x} \mathbb{P}[\hat{Y}=1,X=x| A=0] \nonumber\\  
    & = \sum_{x} \mathbb{P}[\hat{Y}=1|X=x,A=1] \cdot \mathbb{P}[X=x|A=1]- \sum_{x} \mathbb{P}[\hat{Y}=1|X=x,A=0] \cdot \mathbb{P}[X=x|A=0] \nonumber\\
    & \stackrel{(a)}{=} \sum_{x} \hat{Y}_1^x \; \mathbb{P}[X=x| A=1] - \sum_{x} \hat{Y}_0^x \; \mathbb{P}[X=x| A=0] \nonumber\\   
    & \stackrel{(b)}{=} \sum_{x} \hat{Y}_1^x \; \mathbb{P}[X=x] - \sum_{x} \hat{Y}_0^x \; \mathbb{P}[X=x] \nonumber\\ 
  & \stackrel{(c)}{=} \sum_{\substack{x: \\ \Delta^x_1 \geq 0}} \mathbb{P}[X=x] - \sum_{\substack{x: \\ \Delta^x_0 \geq 0}} \mathbb{P}[X=x] 
  \label{Eq:quantity}
\end{align}
In step (a), we replace $\mathbb{P}[\hat{Y}=1|X=x,A=1]$ and $\mathbb{P}[\hat{Y}=1|X=x,A=0]$ by their corresponding abbreviated forms $\hat{Y}_1^x$ and $\hat{Y}_0^x$. Step (b) follows from $X \perp A$. 
Step (c) follows because  $\hat{Y}_1^x = 1$ when $\Delta^x_1 \geq 0$, and $\hat{Y}_1^x = 0$, otherwise. 
Similarly, $ \hat{Y}_0^x = 1$ when  $\Delta^x_0 \geq 0$, and  $\hat{Y}_0^x = 0$, otherwise.

Then, we consider three cases: 
\begin{itemize}
    \item Case $\exists x\; \Gamma^{x}_1 > \Gamma^{x}_0 $. By Assumption~\ref{assumption3} (uniform discrimination) we have that $\forall x \;\Gamma^{x}_1 \geq \Gamma^{x}_0 $.
    Also, recall that $\Gamma^{x}_a \geq 0$ if and only if $\Delta^{x}_a \geq 0$. 
    Therefore, in the expression (\ref{Eq:quantity}), for each $x$ such that 
    $\Delta^{x}_0 \geq 0$, we also have $\Delta^{x}_1 \geq 0$, which concludes the proof for this case.
\item Case $\forall x\; \Gamma^{x}_1 = \Gamma^{x}_0 $. We have that $\Delta^{x}_0 \geq 0$ if and only if $\Delta^{x}_1 \geq 0$, hence the two terms in the expression (~\ref{Eq:quantity}) are equal. 

\item Case $\exists x\; \Gamma^{x}_1 < \Gamma^{x}_0 $. This is the symmetric of the first case. Following the same reasoning (with $0$ and $1$ exchanged), we have that, in the expression (\ref{Eq:quantity}), for each $x$ such that 
    $\Delta^{x}_1 \geq 0$, we also have $\Delta^{x}_0 \geq 0$.
    \end{itemize}
\end{proofE}

Analogously, we have:
\begin{lemmaE}[Quantification of SD$'$][end,restate]
\label{lemma:SD':pre} 
    \begin{align}
    {\rm SD}'\, = \,\begin{cases} 
    \begin{array}{ll}
    \,\mathbb{P}[\Delta'^{X}_1 \geq 0 \; \wedge \; \Delta'^{X}_0 < 0] & \textrm{if} \; \; \exists x \; \Gamma'^{x}_1 > \Gamma'^{x}_0 \\[2mm]    
    \,0 &   \textrm{if} \; \;  \forall x \; \Gamma'^{x}_1 = \Gamma'^{x}_0 \\[2mm]
   \, -  \; \mathbb{P}[\Delta'^{X}_1 < 0 \; \wedge \; \Delta'^{X}_0 \geq 0] &  \textrm{if} \; \; \exists x \; \Gamma'^{x}_1 < \Gamma'^{x}_0 
   \end{array}
    \end{cases} \nonumber
\end{align}
\end{lemmaE}
\begin{proofE}
\begin{align}
    {\rm SD}'  & \stackrel{\rm def}{=} \mathbb{P}[\hat{Y'}=1|A=1] - \mathbb{P}[\hat{Y'}=1|A=0]  \nonumber \\
    & = \sum_{x} \mathbb{P}[\hat{Y'}=1,X=x| A=1] - \sum_{x} \mathbb{P}[\hat{Y'}=1,X=x| A=0] \nonumber\\  
    & = \sum_{x} \mathbb{P}[\hat{Y'}=1|X=x,A=1] \cdot \mathbb{P}[X=x,A=1]- \sum_{x} \mathbb{P}[\hat{Y'}=1|X=x,A=0] \cdot \mathbb{P}[X=x,A=0] \nonumber\\
    & = \sum_{x} \hat{Y'}_1^x \; \mathbb{P}[X=x| A=1] - \sum_{x} \hat{Y'}_0^x \; \mathbb{P}[X=x| A=0] \nonumber\\   
  & {=} \sum_{\substack{x: \\ \Delta'^x_1 \geq 0}} \mathbb{P}[X=x] - \sum_{\substack{x: \\ \Delta'^x_0 \geq 0}} \mathbb{P}[X=x] 
  \nonumber
\end{align}
The proof proceeds like the one in Lemma~\ref{lemma:SD}. We need, however, the following result, which states that 
LDP obfuscation preserves
\textit{uniform discrimination assumption} (Assumption~\ref{assumption3}).

\begin{lemmaE}\label{lemma3}
    If \; 
    $\exists x^* \; \Gamma'^{x^*}_a > \Gamma'^{x^*}_{\overline{a}}$ \; then \; $\forall x  \; \Gamma'^{x}_a \geq \Gamma'^{x}_{\overline{a}}$
\end{lemmaE}
\noindent
{\bf \it Proof}\\
We prove the property by showing that 
$\Gamma'^x_a >  \Gamma'^x_{\overline{a}}$ if and only if 
$\Gamma^x_a >  \Gamma^x_{\overline{a}}$, 
and that
$\Gamma'^x_a =  \Gamma'^x_{\overline{a}}$ if and only if 
$\Gamma^x_a =  \Gamma^x_{\overline{a}}$.
Then, clearly, the statement of the theorem derives from the assumption of \textit{uniform discrimination} for the original data (before obfuscation). 

It is easy to see that
\[
    \Gamma'^{x}_a = \frac{p \Delta^{x}_a + (1-p) \Delta^x_{\overline{a}}}{p \mathbb{P}[X=x,A=a] + (1-p) \mathbb{P}[X=x,A=\overline{a}]} \nonumber
\]

Let us prove that  $\Gamma'^x_a >  \Gamma'^x_{\overline{a}}$ if and only if 
$\Gamma^x_a >  \Gamma^x_{\overline{a}}$:
\[
\begin{array}{rcl}
\Gamma'^x_a &>&  \Gamma'^x_{\overline{a}}
\nonumber \\
    &\Leftrightarrow
    \nonumber
    \\
    \frac{p \Delta^{x}_a + (1-p) \Delta^x_{\overline{a}}}{p \mathbb{P}[X=x,A=a] + (1-p) \mathbb{P}[X=x,A=\overline{a}]}& > & \frac{p \Delta^x_{\overline{a}} + (1-p) \Delta^{x}_a}{p \mathbb{P}[X=x,A=\overline{a}] + (1-p) \mathbb{P}[X=x,A=a]} \nonumber \\
    &\Leftrightarrow
    \nonumber
    \\
    p^2 \Delta^{x}_a \mathbb{P}[X=x,A=\overline{a}] + (1-p)^2 \Delta^x_{\overline{a}} \mathbb{P}[X=x,A=a] & > & p^2 \Delta^x_{\overline{a}} \mathbb{P}[X=x,A=a] + (1-p)^2 \Delta^{x}_a \mathbb{P}[X=x,A=\overline{a}] 
    \nonumber\\
    &\Leftrightarrow   
    \nonumber
    \\
    \left.
    \begin{array}{r}
    p^2 \Gamma ^{x}_a \mathbb{P}[A=a,X=x]\mathbb{P}[A=\overline{a},X=x] \\+ \\
    (1-p)^2 \Gamma^x_{\overline{a}} \mathbb{P}[A=\overline{a},X=x] \mathbb{P}[A=a,X=x]
    \end{array}
    \right\}
    & > & 
    \left\{
    \begin{array}{l}
    p^2 \Gamma^x_{\overline{a}} \mathbb{P}[A=\overline{a},X=x]\mathbb{P}[A=a,X=x] \\+\\
    (1-p)^2 \Gamma^{x}_a \mathbb{P}[A=\overline{a},X=x]\mathbb{P}[A=a,X=x]
    \end{array}
    \right.
    \nonumber \\
    & \Leftrightarrow
    \nonumber
    \\\mathbb{P}[A=\overline{a},X=x]\mathbb{P}[A=a,X=x] \left( p^2 \Gamma ^{x}_a + (1-p)^2 \Gamma^x_{\overline{a}} \right )& > & \mathbb{P}[A=\overline{a},X=x]\mathbb{P}[A=a,X=x] \left( p^2 \Gamma^x_{\overline{a}} + (1-p)^2 \Gamma ^{x}_a \right ) \nonumber \\
     &\Leftrightarrow
    \nonumber
    \\ p^2 \left(\Gamma ^{x}_a - \Gamma^x_{\overline{a}} \right ) & > & (1-p)^2 \left(\Gamma ^{x}_a - \Gamma^x_{\overline{a}} \right ) \nonumber \\
     & \Leftrightarrow
    \nonumber
    \\ \Gamma ^{x}_a & >  & \Gamma^x_{\overline{a}} \nonumber
\end{array}
\]
The property $\Gamma'^x_a =  \Gamma'^x_{\overline{a}}$ if and only if 
$\Gamma^x_a =  \Gamma^x_{\overline{a}}$ can be proved similarly, just replace the ``$>$'' symbol by ``$=$''.   
\end{proofE}

Using Lemma~\ref{lemma1}, by case analysis, the quantification of SD$'$ can be reformulated in terms of the distribution in the original data, as follows.
\begin{lemma}[Quantification of SD$'$ in terms of the distribution on the original data]
\label{lemma:SD'} 
\begin{align}
\footnotesize
    {\rm SD}'\, = \,\begin{cases} 
    \begin{array}{ll}
    \mathbb{P}\left[ \begin{array}{l}
    \Delta^{X}_1 > 0 \; \wedge \; \Delta^{X}_0 < 0 \; \wedge \\
    \; e^{\varepsilon} \geq -\nicefrac{\Delta^{X}_0}{\Delta^{X}_1} \; \wedge \; e^{\varepsilon} > -\nicefrac{\Delta^{X}_1}{\Delta^{X}_0}
    \end{array}
    \right] 
    & \textrm{if} \; \; \exists x \; \Gamma^{x}_1 > \Gamma^{x}_0 
     \\[5mm] 
   0 &\textrm{if} \; \; \forall x \; \Gamma^{x}_1 = \Gamma^{x}_0 
   \\[2mm]
   -  \mathbb{P}\left[
   \begin{array}{l}\Delta^{X}_1 < 0 \; \wedge \; \Delta^{X}_0 > 0 \; \wedge \\
   e^{\varepsilon} > -\nicefrac{\Delta^{X}_0}{\Delta^{X}_1} \; \wedge \; e^{\varepsilon} \geq -\nicefrac{\Delta^{X}_1}{\Delta^{X}_0}
   \end{array}
   \right] 
   &\textrm{if} \; \; \exists x \; \Gamma^{x}_1 < \Gamma^{x}_0 
   \end{array}
    \end{cases} \nonumber
\end{align}
\end{lemma}

We can now state the main result of this section: If $X\perp A$, then, like in the case of conditional statistical disparity, we have that SD$'$ is always sandwiched between SD and $0$.

\begin{thm}[Impact of LDP on SD. Case $X \perp A$]\label{th:sd_indep} \ \\
\begin{enumerate}
\vspace{-0.1cm}
    \item if \; {\rm SD} $> 0$ \;then \;$0 \leq {\rm SD}' \leq {\rm SD}$
    \vspace{0.2cm}
    \item if \; ${\rm SD} < 0$ \;then \;${\rm SD} \leq {\rm SD}' \leq 0$
    \vspace{0.2cm}
    \item if \; ${\rm SD} = 0$\; then \;${\rm SD}' =  {\rm SD} = 0$
\end{enumerate}
\end{thm}
The proof follows immediately from Lemmas~\ref{lemma:SD}  and \ref{lemma:SD'} because the values of $X$ that constitute the probability mass in the expression of SD$'$ are a subset of those that constitute the probability mass in the expression of SD.

\subsubsection*{Discussion}
Theorem~\ref{th:sd_indep} means that, from an unfair situation (SD $>0$ or SD $<0$), obfuscating the sensitive attribute $A$ in general advantages the \textit{unprivileged} group, but it never ends up discriminating the other group. 
(We will see in the next section that this is not always the case when some proxies to the sensitive attribute $A$ exist in the data.)

In one extreme case, the situation does not change (SD$'= $ SD). By looking at the expression quantifying SD and SD$'$ in Lemma~\ref{lemma:SD} and \ref{lemma:SD'}, we can see that this happens when the noise we inject is small, i.e., for high values of $\varepsilon$, and, more precisely, when $\varepsilon$  satisfies $\forall x \,\, \varepsilon \geq \max\{\ln(-\nicefrac{\Delta^{x}_0}{\Delta^{x}_1}), \ln(-\nicefrac{\Delta^{x}_1}{\Delta^{x}_0})\}$. 

In the opposite extreme case, the discrimination is totally eliminated (SD$'= 0$).  
This last case raises when we inject enough noise, and more precisely, when $\varepsilon$ satisfies $\forall x \,\, \varepsilon < \max\{\ln(-\nicefrac{\Delta^{x}_0}{\Delta^{x}_1}), \ln(-\nicefrac{\Delta^{x}_1}{\Delta^{x}_0})\}$. 

In all the other cases, i.e., when for some 
$x$ we have: \\ $\varepsilon \geq \max\{\ln(-\nicefrac{\Delta^{x}_0}{\Delta^{x}_1}), \ln(-\nicefrac{\Delta^{x}_1}{\Delta^{x}_0})\}$ and for other $x$ we have: \\ $\varepsilon < \max\{\ln(-\nicefrac{\Delta^{x}_0}{\Delta^{x}_1}), \ln(-\nicefrac{\Delta^{x}_1}{\Delta^{x}_0})\}$, obfuscation removes some discrimination, but not entirely. Namely $0 < {\rm SD}' < {\rm SD}$ if SD is positive, or ${\rm SD} <  {\rm SD}' < 0 $ if SD is negative.

Note that the extreme case in which $\varepsilon$ is $0$ is equivalent to eliminating $A$ entirely from the data. Hence, the takeout of this section is that the disparity between groups can be eliminated by 
removing the sensitive attribute, but it is important to remember that this is true only because there are no proxies to the sensitive attribute in the data ($X \perp A$). 

Again, we note that Theorem~\ref{th:sd_indep} does not depend on whether the \textit{unprivileged} group is the minority or the majority of the population.  
\\
\noindent
\subsubsection{Second scenario: $X \not\!\perp\!\!\! \; \; A$}\label{subsubsec:dependent_xa}  \ \\
\noindent
Usually, proxy attributes to the sensitive attribute $A$ exist in the data. In other words, $A$ and $X$ are dependent ($X \not\!\perp\!\!\! \; \; A$). In this section, we study the impact of privacy on SD when $X \not\!\perp\!\!\! \; \; A$. Theorem~\ref{th:sd_dep} presents the results of the impact of privacy on SD in this scenario.

\begin{thmE}[Impact of LDP on SD. Case $X \not\!\perp\!\!\! \; \;  A$][end,restate]\ \\
\label{th:sd_dep}
    \begin{enumerate}
    \vspace{-0.1cm}
    \item if \; $\exists x \; \Gamma^{x}_1 > \Gamma^{x}_0$ \; then \; ${\rm SD}' \leq {\rm SD} $ 
    \vspace{0.2cm}
    \item if \; $\exists x \; \Gamma^{x}_1 < \Gamma^{x}_0$ \; then \; ${\rm SD} \leq {\rm SD}' $ 
    \vspace{0.2cm}
    \item if \; $\forall x \; \Gamma^{x}_1 = \Gamma^{x}_0$ \; then \; ${\rm SD}' = {\rm SD}$
    \end{enumerate}
\end{thmE}
\begin{proofE}
We prove the result for the case $\exists x\;\Gamma^{x}_1 > \Gamma^{x}_0$, the other two cases can be proven similarly.\\ 
Recall that:
\begin{align}
    {\rm SD}  & = \sum_{\substack{x: \\ \Delta^x_1 \geq 0}} \mathbb{P}[X=x|A=1] - \sum_{\substack{x: \\ \Delta^x_0 \geq 0}} \mathbb{P}[X=x|A=0] \quad \scriptsize{ \hat{Y}_1^x, \hat{Y}_0^x = 1} \nonumber
\end{align}
Since we are considering the case $\exists x\;\Gamma^{x}_1 > \Gamma^{x}_0$, from   
Assumption~\ref{assumption3} (\textit{uniform discrimination}) we derive $\forall x\;\Gamma^{x}_1 \geq \Gamma^{x}_0$, hence: 
\begin{align}
    {\rm SD}  & = \sum_{\substack{x: \\ \Delta^x_1,\Delta^x_0 \geq 0}} \mathbb{P}[X=x|A=1] - \mathbb{P}[X=x|A=0] \; \; + \sum_{\substack{x: \\ \Delta^x_1 \geq 0, \Delta^x_0 < 0}} \mathbb{P}[X=x|A=1]  \nonumber
\end{align}
After obfuscation, from Lemma~\ref{lemma3} we have that   $\forall x\; \Gamma'^{x}_1 > \Gamma'^{x}_0$. Hence: 
\begin{align}
    {\rm SD}'  & = \sum_{\substack{x: \\ \Delta'^x_1,\Delta'^x_0 \geq 0}} \mathbb{P}[X=x|A=1] - \mathbb{P}[X=x|A=0] \; \; + \sum_{\substack{x: \\ \Delta'^x_1 \geq 0, \Delta'^x_0 < 0}} \mathbb{P}[X=x|A=1]  \nonumber\\
    & = \sum_{\substack{x: \\ \Delta^x_1,\Delta^x_0 \geq 0}} \mathbb{P}[X=x|A=1] - \mathbb{P}[X=x|A=0] \; \; + \sum_{\substack{x: \\ \Delta^x_1 > 0, \Delta^x_0 < 0, \\ - \frac{\Delta^x_0}{\Delta^x_1} \leq e^\varepsilon \leq 
- \frac{\Delta^x_1}{\Delta^x_0}}
    } \mathbb{P}[X=x|A=1] - \mathbb{P}[X=x|A=0]  \nonumber \\ 
     &\quad  +  \sum_{\substack{x: \\ \Delta^x_1 \geq 0, \Delta^x_0 <0\\ e^\varepsilon \geq 
- \frac{\Delta^x_0}{\Delta^x_1}, \\ e^\varepsilon > 
- \frac{\Delta^x_1}{\Delta^x_0}\\}} \mathbb{P}[X=x|A=1]  \nonumber
\end{align}
By case analysis, and similar to the proof of Theorem~\ref{th:sd_indep}, we can conclude Theorem~\ref{th:sd_dep}. The main difference with  Theorem~\ref{th:sd_indep}, is that 
SD$'$ contains the additional  term 
\[ \sum_{\substack{x: \\ \Delta^x_1 > 0, \Delta^x_0 < 0, \\ - \frac{\Delta^x_0}{\Delta^x_1} \leq e^\varepsilon \leq 
- \frac{\Delta^x_1}{\Delta^x_0}}
    } \mathbb{P}[X=x|A=1] - \mathbb{P}[X=x|A=0]
    \]
which can be negative and large enough to cause SD$'$ to go below $0$.
Hence SD$'$ and SD can be of opposite signs.
\end{proofE}

\subsubsection{Discussion}\ \ \ \ 

Theorem~\ref{th:sd_dep} confirms that also in the case $X\not\perp\!\! A$, in general, the \textit{unprivileged} group benefits from privacy, and again, it does not depend on the \textit{privileged} group being the majority or not. This finding is validated by our experiments on both synthetic and real-world datasets (cf. Figures~\ref{fig:results_synthetic_data2} and~\ref{fig:results_real_data2} in Section~\ref{sec:results}).

Theorem~\ref{th:sd_dep} differs from Theorem~\ref{th:sd_indep} mainly on two points. First, SD and SD$'$ can have opposite signs. In other words, from a scenario where there is discrimination against one group, for instance, the group $A=0$ (SD $> 0$), we can have, after obfuscation, a discrimination against the other group $A=1$ (SD$' <0$). We can even have scenarios in which, after obfuscation, the magnitude of unfairness against the other group is higher than the original one. This result is quite surprising.   We simulated such a scenario using synthetic data (S5) and presented the results in Figure~\ref{fig:s5} (Section~\ref{sec:results}).

Second, we note that in case 1 we can have SD $<0$ despite the fact that $\exists x \; \Gamma^{x}_1 > \Gamma^{x}_0$ (which, by the Assumption~\ref{assumption3}, implies that $\forall x \; \Gamma^{x}_1 \geq \Gamma^{x}_0$), and similarly for case 3. From the proof of the above theorem, we can see that it is particularly likely to happen when $\mathbb{P}[X=x | A=1] \ll [X=x | A=0])$. 
This is a form of the \textit{Simpson's paradox} called \textit{Association Reversal}~\cite{pearl2022comment}:  we have a scenario in which for all sub-populations (i.e., for all $x$) there is discrimination against one group, while when considering the whole population, the discrimination is against the other group.
Note that privacy obfuscation does not break the paradox, because also SD$'$ is negative. 

Another form of the \textit{Simpson's paradox} called the \textit{Yule's Association Paradox}~\cite{David2001} can happen when for all sub-populations, the model shows fair results (i.e., $\forall x \; $ CSD$_x =0$), while for the whole population, it shows unfair results (SD$\neq 0$). 
In Section~\ref{sec:results}, we generated a synthetic dataset (S4) to illustrate such a paradox. Note that in this case, the privacy obfuscation has no effect on fairness: all the metrics remain the same. Indeed, if $\forall x \; $ CSD$_x =0$, then   $\forall x \; $ CSD$'_x =0$, and all the metrics under consideration in this paper are based on CSD$'_x$.

\subsection{Impact of LDP on equal opportunity difference} \label{sub-sec:ldp_eod}
This section considers the impact of privacy on EOD (Eq.~\ref{eq:EOD}). This notion of fairness, by contrast to SD (Eq.~\ref{eq:sd}), considers, in addition to the prediction $\hat{Y}$, the true decision $Y$ (cf. Equation~\ref{eq:EOD}).

The justification for the EOD as a notion of fairness is that 
 $Y$ is supposed to be reliable and not to incorporate any bias (Hardt et al.~\cite{hardt2016equality}). Hence, if $\hat{Y}$ is consistent with $Y$, the prediction should be fair as well.  Furthermore, thanks to this compatibility, and in contrast to other notions of fairness, EOD is, in general, going well along with accuracy (although there are exceptions:~\cite{DBLP:conf/aaai/PinzonPPV22} has shown that, for certain distributions, Equal Opportunity implies trivial accuracy).  
 We capture this principle in Assumption~\ref{assumption4} here below, which states that the true decision 
 $Y$ is independent of the sensitive attribute 
 $A$ given $X$.

\begin{assumption}\label{assumption4} Reliable $Y$.
\label{eq:reliable_y}
The decision $Y$ is independent of the sensitive attribute for any value of the explaining variable. Namely: 
\begin{align}
   \mathbb{P}[Y=1 \mid X=x, A=1] = & \; \mathbb{P}[Y=1 \mid X=x, A = 0].\nonumber 
\end{align}
\end{assumption}
The limitation of EOD is that the ``true'' $Y$ may not always be available. 
In its stead, the data may contain decisions that have been made in the past (which may not always have been fair), or decisions based on some proxy to the true $Y$. In any case, Assumption~\ref{assumption4}, may not always be satisfied in the data. 
When it is satisfied, however, we can obtain a strong result about the effect of privacy on EOD, similar to the one for SD. This is expressed by the theorem below.  

\begin{thmE}[Impact of LDP on EOD][end,restate]\ \\
\label{th:eod}
\begin{enumerate}
\vspace{-0.1cm}
    \item if \; ${\rm EOD} > 0$\;  then \; $0 \leq {\rm EOD}' \leq {\rm EOD}$
    \vspace{0.2cm}
    \item if \; ${\rm EOD} < 0$\;  then \; ${\rm EOD} \leq {\rm EOD}' \leq 0$
    \vspace{0.2cm}
    \item if \; ${\rm EOD} = 0$\;  then \; ${\rm EOD}' =  {\rm EOD} = 0$
\end{enumerate}
\end{thmE}
\begin{proofE}
    \begin{align}
    {\rm EOD}  & \stackrel{\rm def}{=} \mathbb{P}[\hat{Y}=1|Y=1, A=1] - \mathbb{P}[\hat{Y}=1|Y=1, A=0]  \nonumber \\
    & = \sum_{x} \mathbb{P}[\hat{Y}=1,X=x| Y=1, A=1] - \sum_{x} \mathbb{P}[\hat{Y}=1,X=x| Y=1, A=0] \nonumber\\  
    & = \sum_{x} \mathbb{P}[\hat{Y}=1|X=x,Y=1,A=1] \cdot \mathbb{P}[X=x|Y=1,A=1]- \sum_{x} \mathbb{P}[\hat{Y}=1|X=x,Y=1,A=0] \cdot \mathbb{P}[X=x|Y=1,A=0] \nonumber\\
    & = \sum_{x} \frac{\mathbb{P}[\hat{Y}=1,Y=1|X=x,A=1]}{\mathbb{P}[Y=1|X=x,A=1]} \cdot \mathbb{P}[X=x|Y=1,A=1] -  \sum_{x} \frac{\mathbb{P}[\hat{Y}=1,Y=1|X=x,A=0]}{\mathbb{P}[Y=1|X=x,A=0]} \cdot \mathbb{P}[X=x|Y=1,A=0] \nonumber\\   
  & \stackrel{\rm (a)}{=} \sum_{\substack{x: \\ \Delta^x_1 \geq 0}} \mathbb{P}[X=x|Y=1,A=1] - \sum_{\substack{x: \\ \Delta^x_0 \geq 0}} \mathbb{P}[X=x|Y=1,A=0]  \nonumber\\
    & \stackrel{\rm (b)}{=} \sum_{\substack{x: \\ \Delta^x_1 \geq 0}} \mathbb{P}[X=x|Y=1] - \sum_{\substack{x: \\ \Delta^x_0 \geq 0}} \mathbb{P}[X=x|Y=1] \nonumber
\end{align}
(a) follows from the fact that  both $\frac{\mathbb{P}[\hat{Y}=1,Y=1|X=x,A=1]}{\mathbb{P}[Y=1|X=x,A=1]} $ and $\frac{\mathbb{P}[\hat{Y}=1,Y=1|X=x,A=0]}{\mathbb{P}[Y=1|X=x,A=0]}$ are equal to $1$ for $x: \Delta^x_1 \geq 0$ and $x: \Delta^x_0 \geq 0$, respectively.
And (b) follows because of the reliability assumption~\ref{assumption4}.\\

Now, after obfuscation and following the same reasoning as in the proofs of Theorems~\ref{th:sd_indep} and ~\ref{th:sd_dep}, we have:
\begin{align}
    {\rm EOD}'  & = \sum_{\substack{x: \\ \Delta^x_1,\Delta^x_0 \geq 0}} \mathbb{P}[X=x|Y=1] - \mathbb{P}[X=x|Y=1] \; \; + \sum_{\substack{x: \\ \Delta^x_1 > 0, \Delta^x_0 < 0, \\ - \frac{\Delta^x_0}{\Delta^x_1} \leq e^\varepsilon \leq 
- \frac{\Delta^x_1}{\Delta^x_0}}
    } \mathbb{P}[X=x|Y=1] - \mathbb{P}[X=x|Y=1]  \nonumber \\ 
     &\quad  +  \sum_{\substack{x: \\ \Delta^x_1,\Delta^x_0 \geq 0, \\ e^\varepsilon \geq 
- \frac{\Delta^x_0}{\Delta^x_1}, \\ e^\varepsilon > 
- \frac{\Delta^x_1}{\Delta^x_0}\\}} \mathbb{P}[X=x|Y=1] - \mathbb{P}[X=x|Y=1]  \nonumber
\end{align}
The rest is deduced by case analysis.

\end{proofE}

We recall that the above theorem holds under Assumption~\ref{assumption4}.  On the other hand, it is valid regardless of whether $X$ and $A$ are independent.


\section{Experimental Results and Discussion}
\label{sec:results}
\subsubsection{Main results and discussion}\label{subsec:results_discussion}

To validate our theoretical results, we have conducted a set of experiments on both synthetic and real-world fairness benchmark datasets. To each of these datasets, the fairness metrics presented in section ~\ref{subsec:fairness} are applied to the baseline model $\mathcal{M}$ (model trained on the original samples) and to the LDP model $\mathcal{M'}$ (model trained on the obfuscated samples). Then, to assess the impact of privacy on fairness, in relation to Theorems  ~\ref{th:csd}, \ref{th:sd_indep}, \ref{th:sd_dep}, and \ref{th:eod}, the predictions of these two models are compared. Recall that the testing samples for both $\mathcal{M}$ and $\mathcal{M'}$ are always kept original without obfuscation. We vary the privacy parameter $\varepsilon$ in the $\{16, 8, 2, 1, 0.85, 0.5, 0.4, 0.3, 0.2,0.1\}$ for the synthetic datasets and in the $\varepsilon = \{16,8,5,4,3,2,1,0.5\}$ for the real-world datasets. At $\varepsilon = 0.1$ (strong privacy), the ratio of probabilities is bounded by $\varepsilon^{0.1} \approx 1.05$, giving nearly indistinguishable distributions between the two groups, whereas at $\varepsilon = 16$ (weak privacy), the distributions are nearly the same as in the original data.

\subsection{Data and Experiments}\label{subsec:data_experiments}
\noindent \textbf{\textit{Environment:}} All the experiments are implemented in Python 3. We use \textit{Random Forest} model~\cite{breiman2001random} for classification with its default hyper-parameters and randomly select $80\%$ as the training set and the remaining $20\%$ as the testing set. For the RR mechanism, we use the implementation in Multi-Freq-LDPy~\cite{arcolezi2022multi}. 

\noindent \textbf{\textit{Stability:}} Since LDP protocols, train/test splitting, and ML algorithms are randomized, we report average results over $100$ runs.


\noindent \textbf{\textit{Datasets:}} We validate our theoretical results with six synthetic datasets (Section~\ref{sub:results_synthetic}) and four real-world datasets (Section~\ref{sub:results_real_world}).

\subsection{Synthetic Datasets}  \label{sub:results_synthetic}
\begin{figure}[t]
\centering
\begin{minipage}{0.3\textwidth}
\centering
\begin{tikzpicture}[>=latex',scale=0.8]

    \node (X) at (2,0) [] {{$X$}};
    \node (A) at (0,-1) [] {$A$};
    \node (Y) at (4,-1) [] {$Y$};

    \path[every node/.style={sloped,anchor=south,auto=false}]
    (X) edge [->] node {} (Y)    
    (A) edge [->] node {} (Y) ;   
\end{tikzpicture}
\\
(a) Causal Graph for S1.
\label{fig:CG1} 
\end{minipage}
\hspace{5mm}
\\[3ex]
\begin{minipage}{0.4\textwidth}
\centering
\begin{tikzpicture}[>=latex',scale=0.8]

    \node (X) at (2,0) [] {{$X$}};
    \node (A) at (0,-1) [] {$A$};
    \node (Y) at (4,-1) [] {$Y$};

    \path[every node/.style={sloped,anchor=south,auto=false}]
    (X) edge [->] node {} (A)    
    (X) edge [->] node {} (Y)    
    (A) edge [->] node {} (Y) ;   
\end{tikzpicture}
\\
(b) Causal Graph for S2 and S6.
\label{fig:CG2} 
\end{minipage}
\\[3ex] 
\begin{minipage}{0.4\textwidth}
\centering
\begin{tikzpicture}[>=latex',scale=0.8]

    \node (X) at (2,0) [] {{$X$}};
    \node (A) at (0,-1) [] {$A$};
    \node (Y) at (4,-1) [] {$Y$};

    \path[every node/.style={sloped,anchor=south,auto=false}]
    (A) edge [->] node {} (X)    
    (X) edge [->] node {} (Y)    
    (A) edge [->] node {} (Y) ;   
\end{tikzpicture}
\\ (c) Causal Graph for S3, S4 and S5.
\label{fig:CG34} 
\end{minipage}
\\[2ex]
\caption{Causal graphs of the Synthetic Datasets.}
\label{fig:synthetic}
\end{figure}
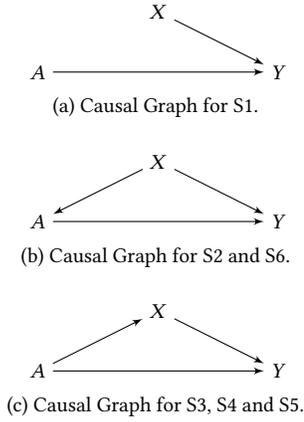

The causal graphs used to generate the synthetic datasets are depicted in Figure~\ref{fig:synthetic}, and the joint empirical probabilities (frequencies) for the various combinations of values are shown in Table~\ref{tab:synth_data_info}. S1 differs from all other datasets in that $X$ and $A$ are independent, whereas in all other datasets, namely S2-S6, $X$ and $A$ are dependent. $A$ and $Y$ are binary variables while $X$ is a discrete variable. In S1, S2, and S4, $X$ is also a binary variable. S3 and S5 are generated to simulate the scenario where privacy shifts discrimination between groups. S5 shows an extreme scenario where $|$SD$'| > |$SD$|$, while $|$SD$'| < |$SD$|$ in S3. And finally, S4 includes a case of \textit{Yule's Association Paradox}~\cite{David2001} (Section~\ref{subsubsec:dependent_xa}).

\begin{table}[t]\label{Table:synthetic-data-probabilities}
    \caption{Distributions of the synthetic datasets.}
 \begin{subtable}{\linewidth}
    \centering
       \caption{S1.}
    \label{tab:s1_data}
    \renewcommand\arraystretch{1.2}
    \begin{tabular}{@{} ll|ll @{}}    \toprule
    \textbf{$Y = 1$}&  & $X=0 $ & $X=1$  \\\midrule
    & $A =1 $& $0.35$ & $0.35$   \\
    & $A =0$ & $0$ & $0.15$   \\
    \hline
    \textbf{$Y = 0$}& & $X=0 $ & $X=1$    \\\hline
    & $A =1$ & $0$ & $0$  \\
    & $A =0$ & $0.15$ & $0$  \\
    \bottomrule
    \hline
 \end{tabular}
  \end{subtable}%
    \vspace{0.2cm}
    \begin{subtable}{\linewidth}
    \centering
    \caption{S2.}
    \label{tab:s2_data}
    \renewcommand\arraystretch{1.2}
    \begin{tabular}{@{} ll|ll @{}}    \toprule
    \textbf{$Y = 1$}&  & $X=0 $ & $X=1$ \\\midrule
    & $A =1 $& $0.28$ & $0.38$ \\
    & $A =0$ & $0$ & $0.12$  \\
    \hline
    \textbf{$Y = 0$}&  & $X=0 $ & $X=1$ \\\hline
    & $A =1$ & $0$ & $0$ \\
    & $A =0$ & $0.22$ & $0$ \\
    \bottomrule
    \hline
    \end{tabular}
    \end{subtable} 
    
     \begin{subtable}{\linewidth}
       \vspace{0.2cm}
\centering
    \caption{S3.}
    \label{tab:s3_data}
    \renewcommand\arraystretch{1.2}
    \begin{tabular}{@{} ll|lll @{}}    \toprule
    \textbf{$Y = 1$}&  & $X=0 $ & $X=1$ & $X=2$  \\\midrule
    & $A =1 $& $0.03$ & $0.17$  &$0.03$ \\
    & $A =0$ & $0$ & $0.17$  & $0.03$ \\
    \hline
    \textbf{$Y = 0$}&  & $X=0 $ & $X=1$ & $X=2$   \\ \hline
    & $A =1$ & $0.24$ & $0.03$  &$0$ \\
    & $A =0$ & $0.1$ & $0.2$  & $0$ \\
    \bottomrule
    \hline
    \end{tabular}
    \end{subtable} 
    \begin{subtable}{\linewidth}
    \vspace{0.2cm}
    \centering
    \caption{S4.}
    \label{tab:s4_data}
    \renewcommand\arraystretch{1.2}
    \begin{tabular}{@{} ll|ll @{}}    \toprule
    \textbf{$Y = 1$}&  & $X=0 $ & $X=1$  \\\midrule
    & $A =1 $& $0$ & $0.4$  \\
    & $A =0$ & $0.03$ & $0.34$  \\
    \hline
    \textbf{$Y = 0$}&   & $X=0 $ & $X=1$   \\ \hline
    & $A =1$ & $0.03$ & $0.07$  \\
    & $A =0$ & $0.13$ & $0$ \\
    \bottomrule
    \hline
    \end{tabular}
    \end{subtable} 
    \begin{subtable}{\linewidth}
    \centering
    \vspace{0.2cm}
    \caption{S5.}
    \label{tab:s5_data}
    \renewcommand\arraystretch{1.2}
    \begin{tabular}{@{} ll|lll @{}}    \toprule
    \textbf{$Y = 1$}&  & $X=0 $ & $X=1$ & $X=2$  \\\midrule
    & $A =1 $& $0.03$ & $0.17$  &$0.03$ \\
    & $A =0$ & $0$ & $0.17$  & $0.03$ \\
    \hline
    \textbf{$Y = 0$}&  & $X=0 $ & $X=1$ & $X=2$  \\\hline
    & $A =1$ & $0.24$ & $0.03$  &$0$ \\
    & $A =0$ & $0.03$ & $0.27$  & $0$ \\
    \bottomrule
    \hline
    \end{tabular}
    \end{subtable} 
    \begin{subtable}{\linewidth}
    \vspace{0.2cm}
    \centering
    \caption{S6.}
    \label{tab:s6_data}
    \renewcommand\arraystretch{1.2}
    \begin{tabular}{@{} ll|lllll @{}}    \toprule
    \textbf{$Y = 1$}&  & $X=0 $ & $X=1$ & $X=2$& $X=3$ & $X=4$  \\\midrule
    & $A =1 $& $0.05$ & $0.08$  &$0.09$&$0.13$&$0.14$ \\
    & $A =0$ & $0.02$ & $0.03$  &$0.06$&$0.03$&$0.04$ \\
    \hline
    \textbf{$Y = 0$}&  & $X=0 $ & $X=1$ & $X=2$& $X=3$ & $X=4$  \\\hline
   & $A =1 $& $0.04$ & $0.02$  &$0.01$&$0.06$&$0$ \\
    & $A =0$ & $0.06$ & $0.04$  &$0.02$&$0.08$&$0$ \\
    \bottomrule
    \hline
    \end{tabular}
    \end{subtable} 
\label{tab:synth_data_info}
\end{table}

\begin{figure}[t]
\centering
    \includegraphics[scale=0.45]{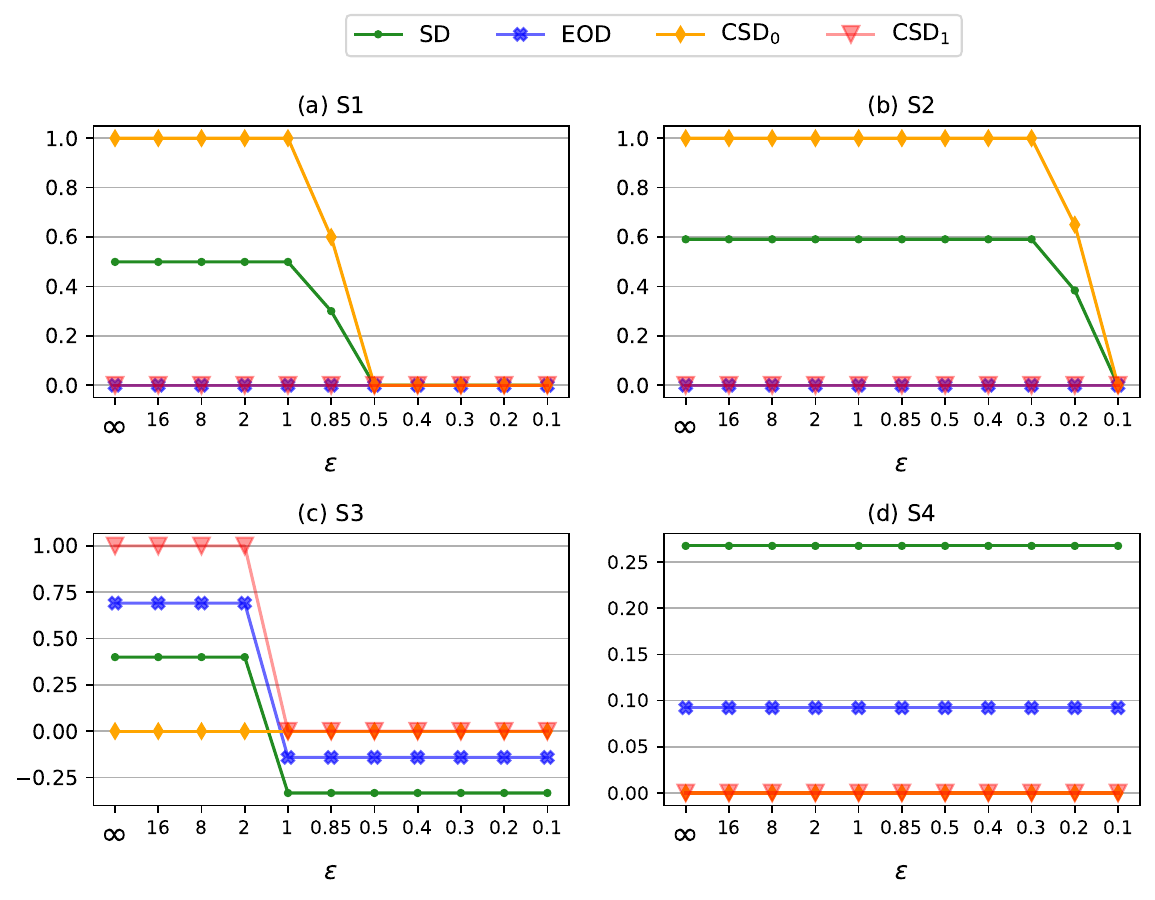} 
    \caption{Results for the synthetic dataset S1-S4, illustrating the impact of LDP on fairness (y-axis) for privacy level $\varepsilon$ (x-axis). 
    Note that in  S3 we have  $X \not\!\perp\!\!\! \; \; A$ and the fairness measure SD is inverted after obfuscation. Also, EOD is inverted after obfuscation. This is because Assumption~\ref{assumption4} is not verified in this dataset. S4 illustrates Yule's Association Paradox, a variant of the  \textit{Simpson's paradox}. The fairness values on the original data (no privacy) are the values for $\varepsilon=\infty$.
}    \label{fig:impact_ldp_fairness_diff_synthetic}
\end{figure}


\begin{figure}[t]
\centering
    \includegraphics[scale=0.4]{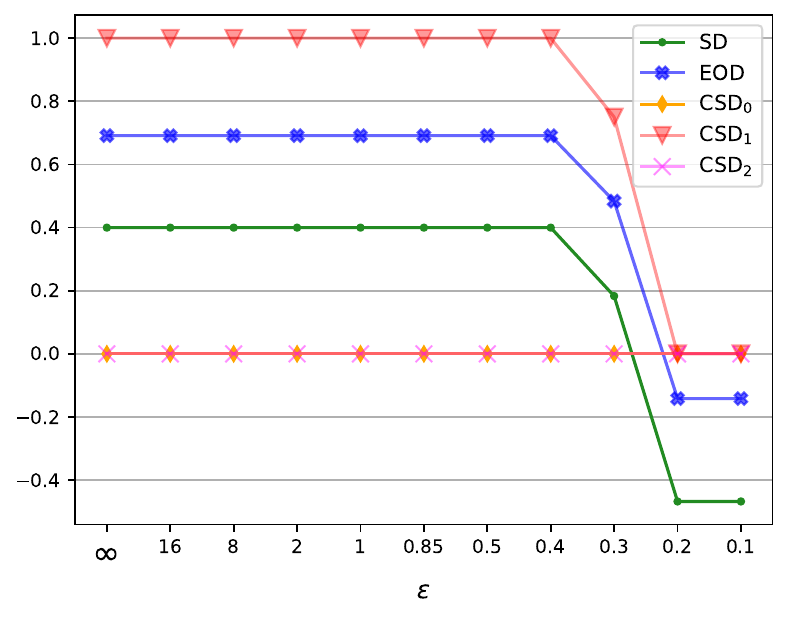} 
    \caption{ Results for the synthetic dataset S5. Note that EOD is also inverted here after obfuscation. Again, this is because Assumption~\ref{assumption4} is not verified in this dataset. }   
    \label{fig:s5}
\end{figure}

\begin{figure}[t]
\centering
    \includegraphics[scale=0.4]{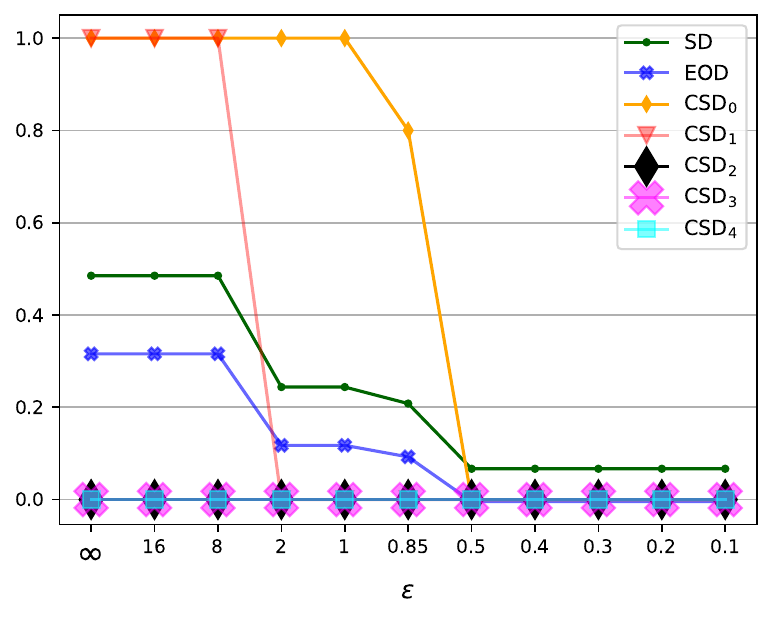} 
    \caption{Results for the synthetic dataset S6. }    \label{fig:s6}
\end{figure}

In the plots that follow, the vertical dashed line in each plot shows the fairness values when the model is trained on original samples (no privacy). It turns out that they are always the same as for the level of privacy $\varepsilon=16$. 

Figure~\ref{fig:impact_ldp_fairness_diff_synthetic} shows the obtained results for S1-S4 presented above while S5 and S6 results are depicted in Figure~\ref{fig:s5} and Figure~\ref{fig:s6}, respectively. 
For example, in S1, where some fairness measures show fair results in the baseline model $\mathcal{M}$, namely EOD and CSD$_{1}$, enforcing privacy helped maintain these fair results: SD$' = 0 $ and CSD$'_{1} = 0$. However, some fairness measures show unfair results against group $A=0$ in the baseline model, namely SD, and CSD$_{0}$; thus, enforcing privacy removed discrimination when enough noise is added. In particular, at $\varepsilon = {\ln(-\nicefrac{\Delta^{x}_0}{\Delta^{x}_1}}) = 0.85$, SD$'$ and CSD$'_{ 0} $ values started to decrease and continued to decrease reaching full parity between groups.

As we proved theoretically in Theorem~\ref{th:sd_dep}, and explained in Section~\ref{subsubsec:dependent_xa}, in S3 and S5 and from a scenario where SD and EOD show discrimination against the group $A=0$, by applying privacy, the discrimination became against the other group $A=1$. Note that this does not contradict Theorem~\ref{th:eod}, because S3 and S5 do not verify Assumption~\ref{assumption4}\footnote{We provide in Appendix~\ref{sec:s7_results} a dataset called S7 that satisfies Assumption~\ref{assumption4}}.
For S3, although this inversion of fairness conclusions (discrimination switching from one group to another when applying privacy), the disparity after obfuscation decreased: $|$SD$'| < |$SD$|$ ($|$SD$'| = 0.33$ and $|$SD$| = 0.39$). However, S5 (Figure~\ref{fig:s5}) shows an extreme case where the disparity between groups after obfuscation increased: $|$SD$'| > |$SD$|$ ($|$SD$'|= 0.46$ and $|$SD$| = 0.39$).  In other words, not only has the discrimination switched from one group to the other after obfuscation, but also the level of unfairness has increased. 

S4 shows a case of the \textit{Yule's Association Paradox}~\cite{David2001}, a variant of the \textit{Simpson's paradox}. That is, the model $\mathcal{M}$ shows fair results for all sub-populations: CSD$_{0} = {\rm CSD}_{1}=0$. However, $\mathcal{M}$ shows unfair results for the whole population: SD$= 0.26$. As shown in Figure~\ref{fig:impact_ldp_fairness_diff_synthetic}(d), the paradox stayed even under a strong privacy regime ($\varepsilon = 0.1$) and hence obfuscating the sensitive attribute solely didn't remove the paradox from the data. 

\begin{figure}
\centering
    \includegraphics[scale=0.4]{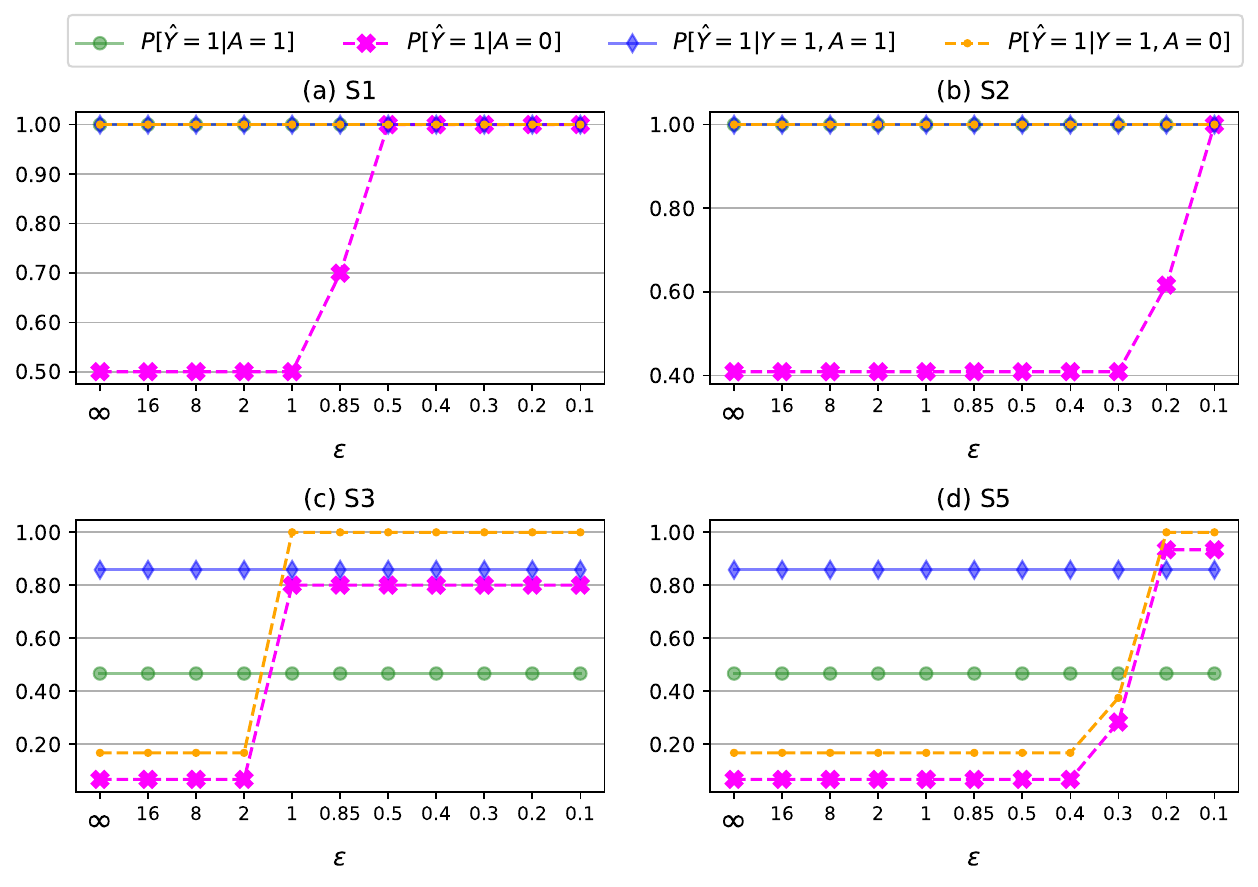} 
    \caption{Impact of LDP on disparity (y-axis) by varying the privacy level $\varepsilon$ (x-axis) showing the behavior of fairness measures on groups separately when applying privacy. For readability, only SD and EOD are illustrated. Results for the synthetic datasets S1, S2, S3, and S5.}
    \label{fig:results_synthetic_data2}
\end{figure}

To better understand how privacy impacts fairness, the plots in Figure~\ref{fig:results_synthetic_data2} show how the impact of privacy on $\mathbb{P}[\hat{Y}=1 \mid A = a]$ and $\mathbb{P}[\hat{Y}=1 \mid Y=1, A = a]$ for both groups $A = 1$ and $A = 0$ while varying $\varepsilon$. 

As mentioned in Section~\ref{subsubsec:dependent_xa}, the \textit{unprivileged} group $A=0$ benefits more from privacy. In other words, when obfuscating the sensitive attribute and aligning with our Theorems~\ref{th:csd}-~\ref{th:eod}, the results of the acceptance rates and the true positive rates of the unprivileged group tend to increase. For instance, for all the synthetic datasets, it is clear that it is group $A=0$ who advantages from privacy as shown in Figure~\ref{fig:results_synthetic_data2}. In other words, there is an increasing trend of $\mathbb{P}[\hat{Y}=1 \mid A = 0]$ and $\mathbb{P}[\hat{Y}=1 \mid Y=1, A = 0]$. 

\subsection{Real-world Datasets} \label{sub:results_real_world}
We consider the following four real-world datasets:

\begin{itemize}
    \item \textit{Compas}: This dataset includes data about defendants from Broward County, Florida, during $2013$ and $2014$ who were subject to \textit{Compas} screening. Various information related to the defendants (e.g., race, {gender, age, arrest date, etc.)} were gathered by ProPublica~\cite{angwin2016machine}, and the goal is to predict a risk score of recidivism\footnote{The risk of recidivism in the \textit{Compas} dataset is a value between $1$ and $10$ where the higher the score, the higher the risk of recidivism for the defendant.}. Only black and white defendants assigned \textit{Compas} risk scores within $30$ days of their arrest are kept for analysis, leading to $5915$ individuals in total. We consider race ($A=1$ for non-black individuals and $A=0$ for black individuals) as the sensitive attribute and the risk of recidivism as the outcome. $Y = 1$ designates a low risk score of recidivism while $Y = 0$ denotes a high risk score. The number of priors of an individual is used as an explaining variable to compute CSD$_x$ where $X=1$ denotes a high number of priors, and $X=0$ denotes a low number of priors.
    
    \item \textit{Adult}~\cite{ding2021retiring}: This dataset~\cite{ding2021retiring} consists of $32,561$ samples, and the goal is to predict the income of individuals based on several personal attributes such as {gender, age, race, marital status, } education, and occupation. Gender is the sensitive attribute ($A=1$ for men and $A=0$ for women), and income is the outcome where $Y = 1$ designates a high income while $Y = 0$ denotes a low income. Education level is the attribute used as an explaining variable to compute CSD$_x$ where $X=1$ denotes a high education level and $X=0$ denotes a low education level.
    
    \item \textit{German credit}~\cite{Dua:2019}: This dataset includes data of $1000$ individuals applying for loans. This dataset is designed for binary classification to predict whether an individual will default on the loan ($Y=0$) or not ($Y=1$) based on personal attributes such as {gender, job,  credit amount,} credit history, etc. We consider gender the sensitive attribute where female applicants ($A=0$) are compared to male applicants ($A=1$). Credit history is the explaining attribute used to compute CSD$_x$ where $X = 1$ denotes an applicant who has duly repaid in the past while $X=0$ denotes a critical account for which the applicant has had late payments and/or defaults in the past.
    
    \item \textit{LSAC}: This dataset originates from the Law School Admissions Council (LSAC) National Bar Passage Study~\cite{wightman1998lsac}. The outcome, denoted as ``pass bar", indicates whether a candidate has successfully passed the bar exam ($Y=1$) or not ($Y=0$). The prediction is based on personal information such as race, {gender, family income, LSAT}, undergraduate GPA score, etc. The sensitive attribute is race ($A=0$ for blacks and $A=1$ for other ethnic groups). The explaining variable is the undergraduate GPA score of an applicant where $X=1$ indicates a high GPA and $X=0$ denotes a low GPA.

\end{itemize}

The real-world datasets' distributions are shown in Table~\ref{Table:probabilities-real-datasets}.
\begin{table}[h]
    \caption{Distributions of the real-world datasets.}\label{Table:probabilities-real-datasets}
    \begin{subtable}{\linewidth}
    \centering
    \caption{Compas.}
    \label{tab:compas_data}
    \renewcommand\arraystretch{1.2}
    \begin{tabular}{@{} ll|ll @{}}    \toprule
    \textbf{$Y = 1$}&  & $X=0 $ & $X=1$  \\\midrule
    & $A =1 $& $0.12$ & $0.03$   \\
    & $A =0$ & $0.06$ & $0.03$   \\
    \hline
    \textbf{$Y = 0$}&  &  &    \\ \hline
    & $A =1$ & $0.15$ & $0.1$  \\
    & $A =0$ & $0.25$ & $0.26$  \\
    \bottomrule
    \hline
    \end{tabular} 
    \end{subtable}%
    \vspace{0.3cm}
    \begin{subtable}{\linewidth}
    \centering
    \caption{Adult.}
    \label{tab:adult_data}
    \renewcommand\arraystretch{1.2}
    \begin{tabular}{@{} ll|ll @{}}    \toprule
    \textbf{$Y = 1$}&  & $X=0 $ & $X=1$ \\\midrule
    & $A =1 $& $0.06$ & $0.53$ \\
    & $A =0$ & $0.02$ &  $0.21$ \\
    \hline
    \textbf{$Y = 0$}&  &  & \\\hline
    & $A =1$ & $0.03$ & $0.06$ \\
    & $A =0$ & $0.02$ & $0.07$ \\
    \bottomrule
    \hline
    \end{tabular}
    \end{subtable} \\
    \begin{subtable}{\linewidth}
         \vspace{0.3cm}
 \centering
    \caption{German credit.}
    \label{tab:german_data}
    \renewcommand\arraystretch{1.2}
    \begin{tabular}{@{} ll|ll @{}}    \toprule
    \textbf{$Y = 1$}&  & $X=0 $ & $X=1$   \\\midrule
    & $A =1 $& $0.23$ & $0.27$  \\
    & $A =0$ & $0.08$ & $0.13$   \\
    \hline
    \textbf{$Y = 0$}&  &  &   \\\hline
    & $A =1$ & $0.06$ & $0.13$   \\
    & $A =0$ &$0.01$  &  $0.09$   \\
    \bottomrule
    \hline
    \end{tabular}
    \end{subtable} 
    \begin{subtable}{\linewidth}
    \centering
       \vspace{0.3cm}
   \caption{LSAC.}
    \label{tab:lsac_data}
    \renewcommand\arraystretch{1.2}
    \begin{tabular}{@{} ll|ll @{}}    \toprule
    \textbf{$Y = 1$}&  & $X=0 $ & $X=1$  \\\midrule
    & $A =1 $& $0.43$ & $0.47$  \\
    & $A =0$ & $0.03$ & $0.01$  \\
    \hline
    \textbf{$Y = 0$}&  &  &   \\\hline
    & $A =1$ & $0.02$ & $0.02$  \\
    & $A =0$ & $0.01$ & $0.01$ \\
    \bottomrule
    \hline
    \end{tabular}
    \end{subtable} 
\label{tab:real_data_info}
\end{table}

\begin{figure}[t]
\centering
    \includegraphics[scale=0.4]{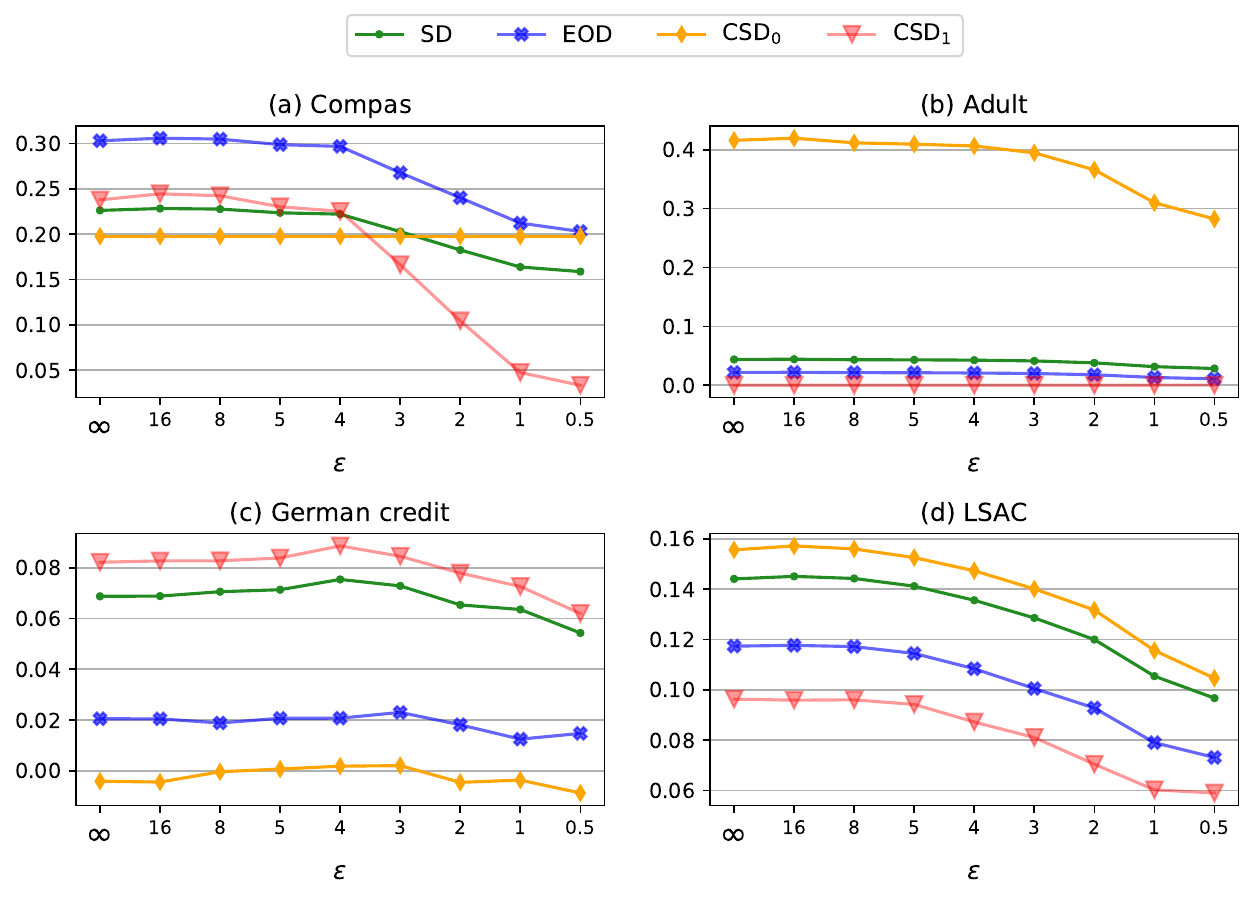} 
    \caption{Results for the real-world datasets. The \emph{German credit} dataset 
    does not satisfy Assumption~\ref{assumption3}, which explains its unstable behavior.}   
    \label{fig:impact_ldp_fairness_diff_real}
\end{figure}

Figure~\ref{fig:impact_ldp_fairness_diff_real} shows the results of applying privacy on the four real-world datasets. As with the synthetic datasets and in alignment with our proofs, obfuscating the sensitive attribute tends to improve the fairness metrics considered in this study in all the datasets {\em except the German credit} one (we will discuss this latter case below). 
We believe that this is due to the fact that the real-world datasets do not always follow the ``ideal'' situation represented by our assumptions. In particular, the datasets we are considering contain other variables besides the $X$ that we use as an explaining variable, which can influence the prediction. 

For instance, in the \textit{Compas} dataset, starting from discrimination against black individuals ($A=0$), privacy reduced the disparity from SD = $0.21$ to $0.15$. Similarly, privacy decreased discrimination against black individuals from SD = $0.13$ to SD = $0.09$ in the \textit{LSAC} dataset, and a similar decrease pattern is observed for all the other fairness measures. We notice a very slight increase of CSD$_1$ at $\varepsilon = 1$, mainly due to ML inaccuracy. The \textit{Adult} dataset also shows a slight disparity decrease caused by privacy. However, starting from a very high disparity between groups given a low level of education (CSD$_0 = 0.39$), the disparity is reduced to $0.22$ at $\varepsilon = 0.1$. 

Concerning the \textit{German credit} dataset, the results show an unstable trend. This is because this data set does not satisfy the \textit{uniform discrimination} assumption (Assumption~\ref{assumption3}). 
Indeed, we for $X=0$, we have, for group $A=1$:
\begin{align}
\Gamma^0_1\quad=\quad \mathbb{P}[Y=1 \mid X = 0, A = 1] - \mathbb{P}[Y=0 \mid X = 0, A = 1]\nonumber 
\\
\quad=\quad
\frac{0.23}{0.29} - \frac{0.06}{0.29}\nonumber \\
\approx \quad
0.58 \nonumber
\end{align}
while for the same $X=0$, for group $A=0$ we have:
\begin{align}
\Gamma^0_0\quad=\quad \mathbb{P}[Y=1 \mid X = 0, A = 0] - \mathbb{P}[Y=0 \mid X = 0, A = 0]\nonumber 
\\
\quad=\quad
\frac{0.08}{0.09} - \frac{0.01}{0.09} \nonumber 
\\
\quad \approx \quad
0.77\nonumber 
\end{align}
Hence $\Gamma^0_1 < \Gamma^0_0$. 

\noindent
On the other hand, for $X=1$ and  group $A=1$ we have:
\begin{align}
\Gamma^1_1\quad=\quad \mathbb{P}[Y=1 \mid X = 1, A = 1] - \mathbb{P}[Y=0 \mid X = 1, A = 1]\nonumber 
\\
\quad=\quad
\frac{0.27}{0.40} - \frac{0.13}{0.40} \nonumber 
\\
\quad \approx \quad
0.35\nonumber 
\end{align}
while for the same $X=1$, for group $A=0$ we have:
\begin{align}
\Gamma^1_0\quad=\quad \mathbb{P}[Y=1 \mid X = 1, A = 0] - \mathbb{P}[Y=0 \mid X = 1, A = 0]\nonumber 
\\
\quad=\quad
\frac{0.13}{0.22} - \frac{0.09}{0.22} \nonumber 
\\
\quad \approx \quad
0.18\nonumber 
\end{align}
Hence, 
$\Gamma^1_1 > \Gamma^1_0$, which means that the \emph{German credit} dataset 
    does not satisfy Assumption~\ref{assumption3} 
It may also mean that the attribute ``Credit history'' is badly chosen as an explaining variable, and\slash or that it is not the main attribute influencing the decision.

\begin{figure}
\centering
    \includegraphics[scale=0.4]{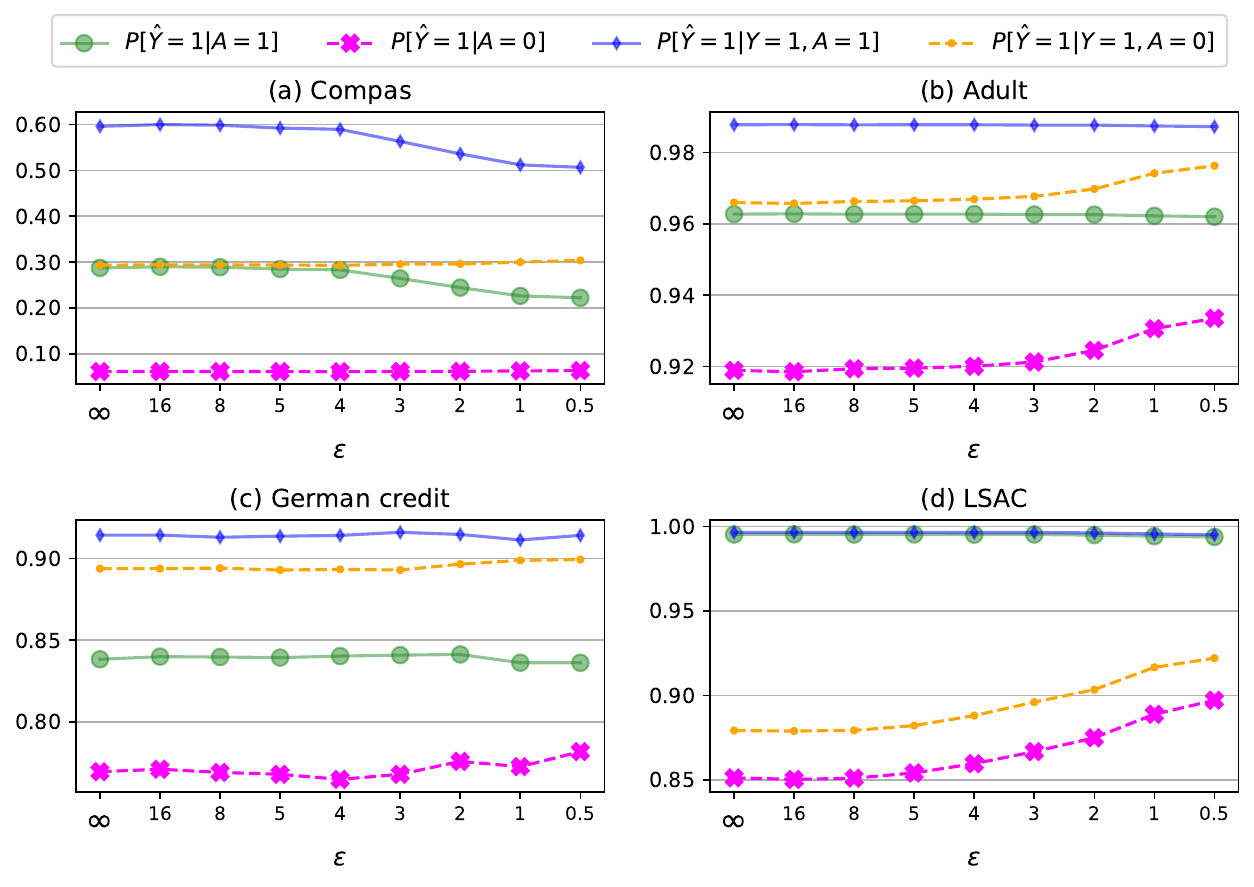} 
    \caption{Impact of LDP on disparity (y-axis) by varying the privacy level $\varepsilon$ (x-axis) showing the behavior of fairness measures on groups separately when applying privacy. For readability, only SD and EOD are illustrated. Results for the real-world datasets.} 
    \label{fig:results_real_data2}
\end{figure}

To better understand how privacy impacts fairness, the plots in Figure~\ref{fig:results_real_data2} show how the impact of privacy on $\mathbb{P}[\hat{Y}=1 \mid A = a]$ and $\mathbb{P}[\hat{Y}=1 \mid Y=1, A = a]$ for both groups $A = 1$ and $A = 0$ while varying $\varepsilon$.

For instance, for the \textit{Adult} dataset, we can observe that women's acceptance rate ($\mathbb{P}[\hat{Y}=1 \mid A = 0]$) and true positive rate increase ($\mathbb{P}[\hat{Y}=1 \mid Y=1, A = 0]$) from $0.91$ to $0.93$ and from $0.96$ to $0.99$, respectively. However, no change is observed for men ($A=1$) even at strong privacy ($\varepsilon = 0.5$). A similar behavior is observed for the \textit{LSAC} dataset. For the \textit{Compas} dataset, while no change is observed for the black defendants' ($A=0$) rates, a decrease is observed for the non-black defendants ($A=1$). Similar behavior is also observed for the \textit{German credit} dataset, where a slight increase in the acceptance rate and the true positive rate for women is observed while almost no change is observed for men.

\subsection{LDP impact on model accuracy}
\begin{figure}[h]
\centering
    \includegraphics[scale=0.5]{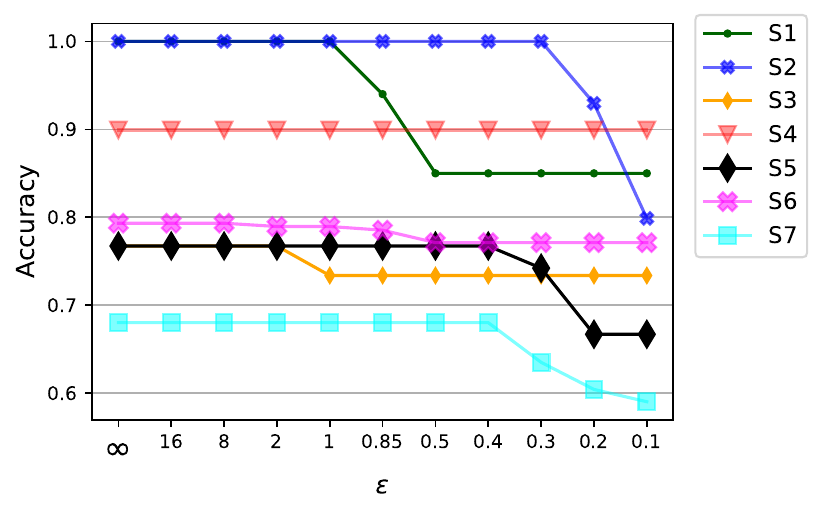} 
    \caption{ Impact of LDP on the model accuracy for the synthetic datasets.}   
    \label{fig:synth_acc}
\end{figure}

\begin{figure}[h]
\centering
    \includegraphics[scale=0.5]{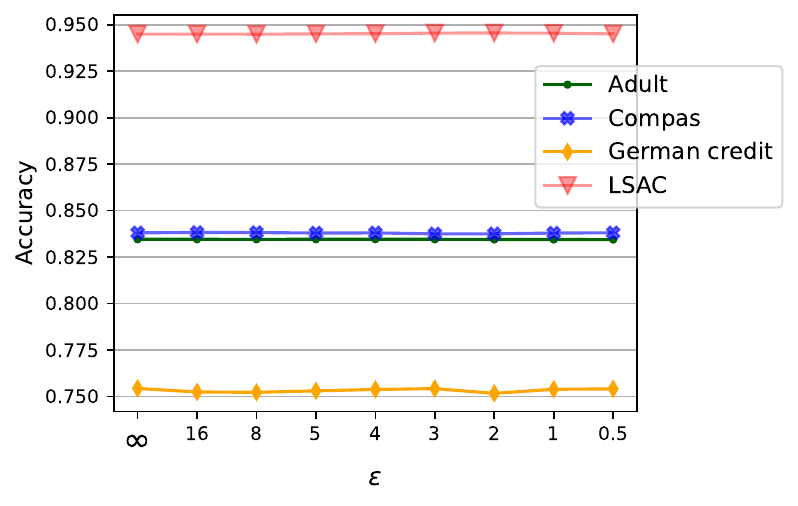} 
    \caption{ Impact of LDP on the model accuracy for the real-world datasets.}   
    \label{fig:real_acc}
\end{figure}

Figures~\ref{fig:synth_acc} and~\ref{fig:real_acc} illustrate the impact of LDP on the accuracy of the model for the synthetic datasets and the real-world datasets, respectively.
From these figures, one can note that, in general, the impact of obfuscating the sensitive attribute on model accuracy of the real-world datasets is minor. The drop in the utility is more apparent for the synthetic datasets but remains reasonable with a maximum drop of $0.2$ in S2.

    

\section{Conclusion}
\label{sec:conclusion}
This study formally examines how LDP affects fairness. More specifically, we provide bounds in terms of the joint distributions and the privacy level, delimiting the extent to which LDP can impact the fairness of the model. Our findings show that the \textit{unprivileged} group benefits more than the \textit{privileged} group when injecting enough noise into the sensitive attribute. Furthermore,  for conditional statistical disparity and for 
equal opportunity difference, injecting noise, in general, improves fairness. 
This also holds for statistical disparity when the data contain no proxies to the sensitive attribute. 
However, when the data contains proxies, in certain cases, by injecting enough noise, while the discrimination was originally against one group, it may be shifted to the other group after obfuscation, and the level of unfairness may be worse than before.
Note that none of our results depend on whether the \textit{unprivileged} group is the minority or the majority.
Additionally, our work focuses on the RR mechanism, a fundamental LDP protocol~\cite{kairouz2016discrete} that serves as a building block for more complex LDP mechanisms (e.g.,~\cite{Bassily2015,tianhao2017,rappor}).

In future work, we aim to extend our work to more fairness measures, particularly overall accuracy equality and others. We also believe that hiding only the sensitive attribute is crucial but not sufficient because proxies for this attribute may exist in the data and thus reveal sensitive information. Therefore, we plan to study formally the impact of LDP on multidimensional data.

\vspace{0.25cm}
\noindent \textbf{Acknowledgement.} This work was partially supported by the European Research Council (ERC) under the European Union’s Horizon 2020 research and innovation programme (Grant agreement 835294) and by the ``ANR 22-PECY-0002'' IPOP (Interdisciplinary Project on Privacy) project of the Cybersecurity PEPR. The work of Héber H. Arcolezi has been partially supported by MIAI @ Grenoble Alpes (``ANR-19-P3IA-0003'').

\bibliographystyle{plain}
\bibliography{ref}

\onecolumn
\appendix

\section{Appendix}
\label{sec:appendix}

\subsection{Proofs}\label{sec:proofs}

\printProofs

\subsection{Results for S7}\label{sec:s7_results}
\begin{figure}[H]
\centering
    \includegraphics[scale=0.45]{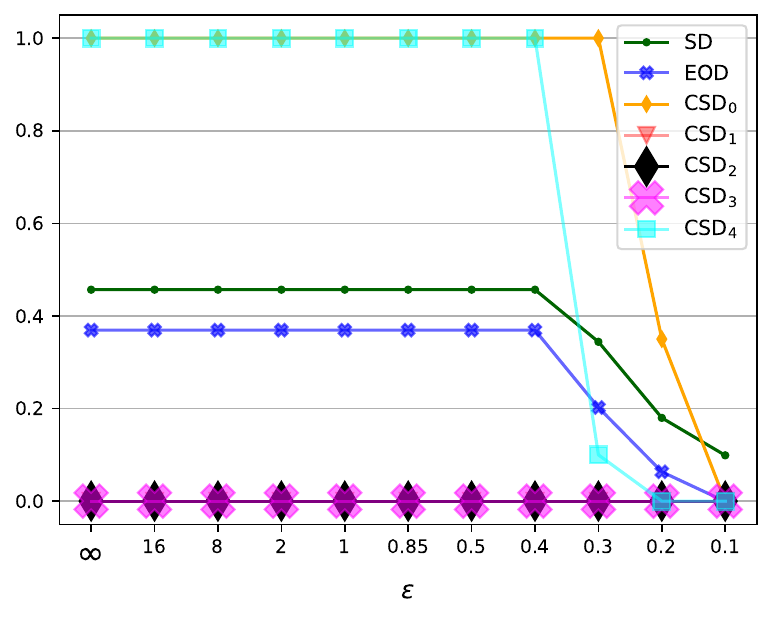} 
    \caption{ Results for the synthetic dataset S7.}   
    \label{fig:s7}
\end{figure}

Below are the data distribution (Table~\ref{tab:s7_data}) and the results of the dataset S7. The data was generated following the causal graph depicted in Figure~\ref{fig:synthetic}(c). The results of applying privacy on fairness are illustrated in Figure~\ref{fig:s7}. Note that in this dataset, the Assumption~\ref{assumption4} is satisfied.
\begin{table}[H]
    \caption{Distributions of the synthetic dataset S7.}
    \centering
    \label{tab:s7_data}
    \renewcommand\arraystretch{1.2}
    \begin{tabular}{@{} ll|lllll @{}}    \toprule
    \textbf{$Y = 1$}&  & $X=0 $ & $X=1$ & $X=2$& $X=3$ & $X=4$  \\\midrule
    & $A =1 $& $0.05$ & $0.07$  &$0.04$&$0.06$&$0.05$ \\
    & $A =0$ & $0.05$ & $0.07$  &$0.04$&$0.06$&$0.05$ \\
    \hline
    \textbf{$Y = 0$}&  & $X=0 $ & $X=1$ & $X=2$& $X=3$ & $X=4$  \\
   & $A =1 $& $0$ & $0.06$  &$0.05$&$0.02$&$0$ \\
    & $A =0$ & $0.09$ & $0.04$  &$0.06$&$0.02$&$0.12$ \\
    \bottomrule
    \hline
    \end{tabular}
\end{table}

\end{document}